\documentclass[generic]{imsart}
\usepackage{amsthm}
\usepackage{amsmath}
\usepackage{natbib}
\usepackage[colorlinks,citecolor=blue,urlcolor=blue,filecolor=blue,backref=page]{hyperref}
\usepackage{graphicx}
\usepackage{cancel}
\usepackage{subfig}
\usepackage{epstopdf}
\usepackage{geometry}
\usepackage{graphics}
\usepackage{booktabs}

\startlocaldefs
\newcommand{\given}{\,|\,}
\newcommand{\T}{\top}

\newcommand{\calS}{{\cal S}}

\newcommand{\calL}{{\cal L}}

\newcommand{\calT}{{\cal T}}
\newcommand{\calU}{{\cal U}}
\newcommand{\calV}{{\cal V}}

\newcommand{\tildew}{\tilde{w}}

\endlocaldefs

\begin{document}

\begin{frontmatter}
 
 \title{Modeling Massive Spatial Datasets Using a Conjugate Bayesian Linear Regression Framework}

 \runtitle{Massive Spatial Data Modeling Using Bayesian Linear Regression}

\begin{aug}
%\author{\fnms{Sudipto} \snm{Banerjee}}
\author{\fnms{Sudipto} \snm{Banerjee}\ead[label=e1]{sudipto@ucla.edu}}
%\and
%\author{\fnms{} \snm{}}

\runauthor{Sudipto Banerjee}

\address[]{UCLA Department of Biostatistics\\ 650 Charles E. Young Drive South\\ Los Angeles, CA 90095-1772.}

%\thankstext{<id>}{<text>}

\end{aug}

\begin{abstract}
Geographic Information Systems (GIS) and related technologies have generated substantial interest among statisticians with regard to scalable methodologies for analyzing large spatial datasets. A variety of scalable spatial process models have been proposed that can be easily embedded within a hierarchical modeling framework to carry out Bayesian inference. While the focus of statistical research has mostly been directed toward innovative and more complex model development, relatively limited attention has been accorded to approaches for easily implementable scalable hierarchical models for the practicing scientist or spatial analyst. This article discusses how point-referenced spatial process models can be cast as a conjugate Bayesian linear regression that can rapidly deliver inference on spatial processes. The approach allows exact sampling directly (avoids iterative algorithms such as Markov chain Monte Carlo) from the joint posterior distribution of regression parameters, the latent process and the predictive random variables, and can be easily implemented on statistical programming environments such as \texttt{R}.
\end{abstract}

%% ** Keywords **

\end{frontmatter}

\section{Introduction}\label{Sec: Introduction}
\noindent Statistical modeling and analysis for spatial and spatial-temporal data continue to receive much attention due to enhancements in computerized Geographic Information Systems (GIS) and accompanying technologies. Bayesian hierarchical spatiotemporal process models have become widely deployed statistical tools for researchers to better understand the complex nature of spatial and temporal
variability
%Two base units of geographic mapping are commonly encountered: locations that are areas or regions with well-defined neighbors (such as pixels in a lattice, counties in a map, etc.), whence they are called \emph{areally referenced} data; or locations that are points with coordinates (latitude-longitude, Easting-Northing etc.), in which case they are called \emph{point referenced} or \emph{geostatistical}. Statistical methods for analyzing point-referenced spatial and spatiotemporal data constitute the subject of geostatistics. 
See, for example, the books \cite{cres93}, \cite{stein99}, \cite{moll03}, \cite{scha04}, \cite{geldigfuegut}, \cite{creswikle11} and \cite{banerjee2014hierarchical} for a variety of statistical methods in diverse applications domains.  

Spatial data analysis is conveniently carried out by embedding a spatial process within the familiar hierarchical modeling paradigm,
\begin{equation}\label{eq: generic_paradigm}
[\mbox{data}\given \mbox{process}] \times [\mbox{process}\given \mbox{parameters}] \times [\mbox{parameters}]\; .
\end{equation}
Modeling for point-referenced data, which refers to data referenced by points with coordinates (latitude-longitude, Easting-Northing etc.), proceeds from a random field that introduces dependence among any finite collection of random variables. Formally, the random field is a stochastic process defined as an uncountable set of random variables, say $\{w(\ell): \ell\in \calL\}$, over a domain of interest $\calL$. This uncountable set is endowed with a probability law specifying the joint distribution for any finite subset of random variables. Spatial processes are usually constructed assuming $\calL\subseteq \Re^d$ (usually $d=2$ or $3$) or, perhaps, as a subset of points on a sphere or ellipsoid. In spatiotemporal settings $\calL = \calS\times \calT$, where $\calS \subset \Re^d$ and $\calT \subset [0,\infty)$ are the space and time domains, respectively, and $\ell = (s,t)$ is a space-time coordinate with spatial location $s \in \calS$ and time point $t\in \calT$ \citep[see, e.g.,][for details]{gnei10}. 

Gaussian random fields are specified with a covariance function $\mbox{cov}\{w(\ell), w(\ell')\} = K_{\theta}(\ell, \ell')$ for any two points $\ell$ and $\ell'$ in $\calL$. If $\calU$ and $\calV$ are finite sets comprising $n$ and $m$ points in $\calL$, respectively, then $K_{\theta}(\calU,\calV)$ denotes the $n\times m$ matrix whose $(i,j)$-th element is evaluated using the covariance function $K_{\theta}(\cdot, \cdot)$ between the $i$-th point $\cal U$ and the $j$-th point in $\calV$. If $\calU$ or $\calV$ comprises a single point, $K_{\theta}(\calU,\calV)$ is a row or column vector, respectively. A valid spatiotemporal covariance function ensures that $K_{\theta}(\calU,\calU)$ is positive definite for any finite set $\calU$, which we will denote simply as $K_{\theta}$ if the context is clear. A customary specification models $\{w(\ell): \ell\in \calL\}$ as a zero-centered Gaussian process, denoted as $w(\ell)\sim GP(0, K_{\theta}(\cdot,\cdot))$. For any finite collection $\calU=\{ \ell_1, \ell_2, \ldots, \ell_n\}$ in $\calL$, the $n\times 1$ random vector $w_{\calU}=(w(\ell_1)), w(\ell_2), \ldots, w(\ell_n))^{\T}$ is distributed as $N(0, K_{\theta})$, where $K_{\theta} = K_{\theta}(\calU,\calU)$. Further details on valid spatial (and spatiotemporal) covariance functions can be found in \cite{gnei10}, \cite{cres93}, \cite{stein99}, \cite{geldigfuegut}, \cite{creswikle11} and \cite{banerjee2014hierarchical} and numerous references therein. %The more common assumptions are of \emph{stationarity} and \emph{isotropy}. The former assumes that $K_{\theta}(\ell,\ell')=K_{\theta}(\ell-\ell')$ depends upon the coordinates only through their separation vector, while isotropy goes a step further and assumes the covariance is a function of the distance between them.  

%Spatial and spatiotemporal processes are conveniently embedded within Bayesian hierarchical models.
If $y(\ell)$ represents a variable of interest at point $\ell$, then a customary spatial regression model at $\ell$ is
\begin{equation}\label{eq: BasicModel}
y(\ell) = x^{\T}(\ell)\beta + w(\ell) + \epsilon(\ell)\; , 
\end{equation}
where $x(\ell)$ is a $p\times 1$ ($p<n$) vector of spatially referenced predictors, $\beta$ is the $p\times 1$ vector of slopes, and $w(\ell)\sim GP(0, K_{\theta}(\cdot,\cdot))$ is the spatial or spatiotemporal process and $\epsilon(\ell)$ is a white noise process modeling measurement error or fine scale variation attributed to disturbances at distances smaller than the minimum observed separations in space and/or time. We now embed (\ref{eq: BasicModel}) and the spatial process within the Bayesian hierarchical model
\begin{align}\label{eq: Bayesian_Spatial_Gaussian_Generic}
p(\theta, \beta, \tau) \times N(w\given 0, K_{\theta}) \times N(y \given X\beta + w, D_{\tau})\; , 
\end{align}
where $y = (y(\ell_1),y(\ell_2),\ldots,y(\ell_n))^{\T}$ is the $n\times 1$ vector of observed outcomes, $X$ is the $n\times p$ matrix of regressors with $i$-th row $x^{\T}(\ell_i)$ and $D_{\tau}$ is the covariance matrix for $\epsilon(\ell)$ over $\{\ell_1,\ell_2,\ldots,\ell_n\}$. A common specification is $D_{\tau} = \tau^2I_n$, where $\tau^2$ is called the ``nugget.'' The hierarchy is completed by assigning prior distributions to $\beta$, $\theta$ and $\tau$. 

For fitting (\ref{eq: Bayesian_Spatial_Gaussian_Generic}) to large spatial datasets, a substantial computational expense is incurred from the size of $K_{\theta}$. Since $\theta$ is unknown, each iteration of the model fitting algorithm will involve decomposing  or factorizing $K_{\theta}$, which typically requires $\sim n^3$ floating point operations (flops) and order of $\sim n^2$ for memory requirements. In geostatistical settings, data are almost never observed on regular grids and the configuration of points are typically highly irregular. The covariance models that have been demonstrated to be most effective for inference do not, in general, result in any computationally exploitable structure for $K_{\theta}$, which makes the matrix computations prohibitive for large values of $n$. For Gaussian likelihoods, one can integrate out the random effects $w$ from (\ref{eq: Bayesian_Spatial_Gaussian_Generic}) and work with the posterior 
\begin{align}\label{eq: Bayesian_Spatial_Gaussian_Generic_Marginalized}
p(\theta, \beta, \tau \given y) \propto p(\theta, \beta, \tau) \times N(y \given X\beta, K_{\theta} + D_{\tau})\; .
\end{align}
This reduces the parameter space to $\{\tau^2,\theta,\beta\}$ by excluding the high-dimensional vector $w$, but one still needs to work with $K_{\theta} + D_{\tau}$, which is $n\times n$. These are referred to as ``big-n'' or ``high-dimensional'' problems in geostatistics.

There is already a substantial literature on high-dimensional spatial and spatiotemporal modeling and we do not attempt to undertake a comprehensive review here; see, e.g., \cite{banerjee2017high} for a focused review on some popular Bayesian approaches and \cite{heatoncontest2019} for a comparative evaluation for a variety of contemporary statistical methods. These papers, and the references therein, offer a variety of algorithmic and model-based approaches for large data. Some published methods have scalable implementations into the millions \citep[see, e.g.,][]{katzfussmultires,abdulah2018,huangsun2018,finley2019efficient,zdb2019} but often require specialized high-performance computer architectures and libraries harnessing parallel processing or graphical processing units. Also, uncertainty quantification on the spatial process while maintaining fidelity to the underlying probability model may also be challenging. With the advent of a new generation of data products, there is a need for some simpler implementations that can be run on modest computing architectures by practicing spatial analysts. This requires new directions in thinking about high-dimensional spatial problems. Here, we will show how some elementary conjugate Bayesian linear regression models can be exploited to conduct Bayesian analysis for massive spatial datasets. While a common underlying idea is to approximate the underlying spatial process with a scalable alternative, we will ensure that such approximations will result in well-defined probability models. In this sense, these approaches can be described as model-based solutions for very large spatial datasets that can be executed on modest computing environments. One exception to the fully model-based approach will be a divide and conquer approach that we briefly review, where an approximation to the full posterior distribution for the entire data is constructed from several posterior distributions of smaller subsets of the data.     

The balance of the paper proceeds as follows. The next section briefly reviews dimension reduction and sparsity inducing spatial models. Section~\ref{sec:conj_bayes_massive_data} presents some standard distribution theory for Bayesian linear regression and outlines how scalable spatial process models can be cast into such frameworks. A synposis of some simulation experiments and data analysis examples are provided in Section~\ref{sec: illustrative_examples}.  Section~\ref{sec: meta-kriging} presents an alternative approach based upon dividing and conquering the data, known as meta-kriging. The paper concludes with some further discussion in Section~\ref{sec: discussion}.  

\section{Dimension reduction and sparsity}\label{sec: dim_reduction_sparsity}

\noindent Dimension reduction \citep[][]{wikle99} is among the most conspicuous of approaches for handling large spatial datasets. This customarily proceeds by representing or approximating the spatial process in terms of the realizations of a latent process over a smaller set of points, often referred to as \emph{knots}. Thus,
\begin{equation}\label{eq:generic_low_rank}
w(\ell) \approx \tilde{w}(\ell) = \sum_{j=1}^{r} b_{\theta}(\ell,\ell_{j}^{*}) z(\ell_{j}^{*}) = b_{\theta}^{\T}(\ell)z,
\end{equation}
where $z(\ell)$ is a well-defined (usually unobserved) process and $b_{\theta}(\cdot,\cdot)$ is a family of basis functions or kernels, possibly depending upon some parameters $\theta$. The collection of $r$ locations $\{\ell_1^*,\ell^*_2,\ldots,\ell^*_r\}$ are the knots, $b_{\theta}(\ell)$ and $z$ are $r\times 1$ vectors with components $b_{\theta}(\ell,\ell_{j}^{*})$ and $z(\ell_j^*)$, respectively. Therefore, $\tildew = B_{\theta}z$, where $\tilde{w} = (\tilde{w}(\ell_1), \tilde{w}(\ell_2),\ldots,\tilde{w}(\ell_{n}))^{\T}$ and $B_{\theta}$ is $n\times r$ with $(i,j)$-th element $b_{\theta}(\ell_i,\ell_{j}^{*})$. We work with $r$ (instead of $n$) $z(\ell_j^*)$'s and the  $n\times r$ matrix $B_{\theta}$. Choosing $r << n$ effectuates dimension reduction because $\tilde{w}(\ell)$, as defined in (\ref{eq:generic_low_rank}), spans only an $r$-dimensional space. When $n>r$, the joint distribution of ${\tilde{w}}$ is singular. Nevertheless, we construct a valid stochastic process with covariance function
\begin{equation}\label{eq:low_rank_corr}
\mbox{cov}(\tilde{w}(\ell), \tilde{w}(\ell')) = b_{\theta}^{\T}(\ell)V_{z}b_{\theta}(\ell')\; ,
\end{equation}
where $V_z$ is the variance-covariance matrix (also depends upon parameter $\theta$) for $z$. From (\ref{eq:low_rank_corr}), we see that, even if $b_{\theta}(\cdot,\cdot)$ is stationary, the induced covariance function is not. If the $z$'s are Gaussian, then $\tilde{w}(\ell)$ is a Gaussian process. Every choice of basis functions yields a process and there are too many choices to enumerate here. Wikle \cite{wikle_2011} offers an excellent overview of low rank models.

Some choices of basis functions can be more computationally efficient than others depending upon the specific application. For example, \cite{cres08} (also see \cite{shicres07}) discuss ``Fixed Rank Kriging'' (FRK) by constructing $B_{\theta}$ using very flexible families of non-stationary covariance functions to carry out high-dimensional kriging, \cite{cres10} extend FRK to spatiotemporal settings calling the procedure ``Fixed Rank Filtering'' (FRF), \cite{katz12} provide efficient constructions for $B_{\theta}$ for massive spatiotemporal datasets, and \cite{katzfuss2013} uses spatial basis functions to capture medium to long range dependence and tapers the residual $w(\ell) - \tildew(\ell)$ to capture fine scale dependence. Multiresolution basis functions \citep{Nychka_Wikle_Royle_2002,nychka2015} have been shown to be effective in building computationally efficient nonstationary models. These papers amply demonstrate the versatility of low-rank approaches using different basis functions. An alternative approach specifies $z(\ell)$ itself as a spatial process. This process is called the ``parent process'' and one can derive a low-rank process $\tilde{w}(\ell)$ from the parent. One such derivation emerges from truncating the Karhunen-Lo\`eve (infinite) basis expansion for a Gaussian process to a finite number of terms to obtain a low-rank process \citep[see, e.g.,][]{rasm08,banerjee2014hierarchical}. This is equivalent to projecting the parent process on a lower-dimensional subspace determined by a partial realization of the parent over $r$ knots of the process. This yields the \emph{predictive process} and several variants aimed at improving the approximation \citep{ban08,ban10,sang11,sang12,katzfussmultires}; also see \citep{finbangel15} and \cite{banerjee2017high} for computational details on efficiently implementing Gaussian predictive processes.

While dimension reduction methods have been applied extensively and effectively to analyze spatial data sets in the order of $n\sim 10^4$, their computational efficiency and inferential performance tend to struggle at even larger scales \citep{banerjee2017high}. More recently, there has been substantial developments in full rank models that exploit sparsity. %Covariance tapering \citep[][]{fur06,kauf08} and Markov random field approximations are examples. What we discuss below, however, are fully process-based models that can interpolate random fields at arbitrary resolutions. 
We introduce sparsity either in the covariance matrix or its inverse (the precision matrix). Covariance tapering \citep{fur06,kauf08,du09} is in the spirit of the former by modeling $\mbox{var}\{w\} = K_{\theta} \odot K_{\mbox{tap},\nu}$, where $K_{\mbox{tap},\nu}$ is a sparse covariance matrix formed from a compactly supported, or \emph{tapered}, covariance function with tapering parameter $\nu$ and $\odot$ denotes the element wise (or Hadamard) product of two matrices. The Hadamard product retains positive definiteness, so $K_{\theta} \odot K_{\mbox{tap},\nu}$ is positive definite. Furthermore, $K_{\mbox{tap},\nu}$ is sparse because a tapered covariance function is equal to $0$ for all pairs of locations separated by a distance beyond a threshold $\nu$. Covariance tapering is undoubtedly an attractive approach for constructing sparse covariance matrices, but its practical implementation for full Bayesian inference will generally require efficient sparse Cholesky decompositions, numerically stable determinant computations and, perhaps most importantly, effective memory management. These issues are yet to be tested for truly massive spatiotemporal datasets with $n \sim 10^5$ or more.

One could also devise models with sparse precision matrices. For finite-dimensional distributions conditional and simultaneous autoregressive (CAR and SAR) models \citep[see, e.g.,][and references therein]{cres93,banerjee2014hierarchical} adopt this approach for areally referenced datasets. The CAR models are special instances of Gaussian Markov random fields or GMRFs \citep{rueheld04} that have led to the popular quadrature based Integrated Nested Laplace Approximation (INLA) algorithms \cite{ruemartinochopin2009} for Bayesian inference and to the approximation of Gaussian  processes\cite{lindgrenruelindstrom2011}. %Thus, a Gaussian process model with a dense covariance function is approximated by a GMRF with a sparse precision matrix. 
These approaches can be computationally efficient for certain classes of covariance functions with stochastic partial differential equations (SPDE) representations (including the versatile Mat\'ern class), but their inferential performance on spatiotemporal or multivariate Gaussian processes (perhaps specified through more general covariance or cross-covariance functions) embedded within Bayesian hierarchical models is yet to be fully developed or assessed for massive datasets.
 
One could also construct massively scalable sparsity-inducing Gaussian processes using essentially the techniques used in graphical Gaussian models by exploiting the relationship between the Cholesky decomposition of a positive definite matrix and conditional independence. For Gaussian processes in particular, recent developments on the Nearest Neighbor Gaussian Processes (NNGP) \citep{datta16,datta16b,banerjee2017high,finley2019efficient} have proceeded from GP likelihoods using directed acyclic graphs (or DAGs) as used by Vecchia \cite{ve88} and Stein et al.\cite{stein04}. The NNGP is a Gaussian process whose finite-dimensional realizations will have sparse precision matrices. Other related papers using the approximation in \cite{ve88} include \cite{stroud17}, \cite{guinness18}, \cite{katzfuss2017general}, and \cite{katzfuss2018predictions}. Shi et al. \cite{Shi2017} recently used the NNGP for uncertainty quantification and Ma et al. \cite{Ma2017FusedGP} used it as a part of a rich class of fused Gaussian process models.

Full Bayesian inference for low-rank and sparse Gaussian process models require iterative algorithms such as Markov chain Monte Carlo (MCMC) or INLA. Details of these implementations can be found in the aforementioned references. In the following section, we will discuss how these spatial models can be embedded within a Bayesian linear regression framework and provide some practical strategies for inference based upon direct (exact) sampling from the posterior distribution.  

\section{Conjugate Bayesian models for massive datasets}\label{sec:conj_bayes_massive_data}
\subsection{Conjugate Bayesian linear geostatistical models}\label{sec: conj_bayes_lm}
\noindent A conjugate Bayesian linear regression model is written as
\begin{align}\label{eq: Conjugate_Bayesian_LM}
 y \given \beta, \sigma^2 &\sim N(X\beta, \sigma^2 V_y)\;;\quad \beta \given \sigma^2 \sim N(\beta \given \mu_{\beta}, \sigma^2 V_{\beta})\;;\quad \sigma^2 \sim IG(a_{\sigma}, b_{\sigma})\; , 
\end{align}
where $y$ is an $n\times 1$ vector of observations of the dependent variable, $X$ is an $n\times p$ matrix (assumed to be of rank $p$) of independent variables (covariates or predictors) and its first column is usually taken to be the intercept, $V_{y}$ is a fixed (i.e., known) $n\times n$ positive definite matrix, $\mu_{\beta}$, $V_{\beta}$, $a_{\sigma}$ and $b_{\sigma}$ are assumed to be fixed hyper-parameters specifying the prior distributions on the regression slopes $\beta$ and the scale $\sigma^2$. This model is easily tractable and the posterior distribution is 
\begin{align}\label{eq: Conjugate_Bayesian_LM_Posterior}
 p(\beta, \sigma^2 \given y) &= \underbrace{IG(\sigma^2\given a^*_{\sigma}, b^*_{\sigma})}_{p(\sigma^2\given y)} \times \underbrace{N(\beta\given Mm, \sigma^2 M)}_{p(\beta\given \sigma^2, y)} \; , 
\end{align}
where $a^*_{\sigma} = a_{\sigma}+n/2$, $b^*_{\sigma} = b_{\sigma} + (1/2)\left\{\mu_{\beta}^{\top}V_{\beta}^{-1}\mu_{\beta} + y^{\top}V_y^{-1}y - m^{\top}Mm\right\}$, $M^{-1} = V_{\beta}^{-1} + X^{\top}V_y^{-1}X$ and $m = V_{\beta}^{-1}\mu_{\beta} + X^{\top}V_y^{-1}y$. Sampling from the joint posterior distribution of $\{\beta,\sigma^2\}$ is achieved by first sampling $\sigma^2 \sim IG(a^*_{\sigma}, b^*_{\sigma})$ and then sampling $\beta \sim N(Mm, \sigma^2 M)$ for each sampled $\sigma^2$. This yields marginal posterior samples from $p(\beta\given y)$, which is a non-central multivariate $t$ distribution but we do not need to work with its complicated density function. See \cite{gelman2013} for further details on the conjugate Bayesian linear regression model and sampling from its posterior. 

We will adapt (\ref{eq: Conjugate_Bayesian_LM}) to accommodate (\ref{eq: Bayesian_Spatial_Gaussian_Generic}) or (\ref{eq: Bayesian_Spatial_Gaussian_Generic_Marginalized}). Let us first consider (\ref{eq: Bayesian_Spatial_Gaussian_Generic_Marginalized}) with the customary specification $D_{\tau} = \tau^2 I$ and let $K_{\theta} = \sigma^2 R(\phi)$, where $R(\phi)$ is a correlation matrix whose entries are given by a correlation function $\rho(\phi;\;\ell_i,\ell_j)$. Thus, $\theta = \{\sigma^2,\phi\}$, where $\sigma^2$ is the spatial variance component and $\phi$ is a spatial decay parameter controlling the rate at which the spatial correlation decays with separation between points. A simple example is $\rho(\phi;\; \ell_i,\ell_j) = \exp(-\phi\|\ell_i-\ell_j\|)$, although much richer choices are available \citep[see, e.g., Ch~3 in][]{banerjee2014hierarchical}. Therefore, we can write $K_{\theta} = \sigma^2 V_y$, where $V_y = R(\phi) + \delta^2I$ and $\delta^2 = \tau^2/\sigma^2$ is the ratio between the ``noise'' variance and ``spatial'' variance. If we assume that $\phi$ and $\delta^2$ are fixed and that the prior on $\{\beta,\sigma^2\}$ are as in (\ref{eq: Conjugate_Bayesian_LM}), then we have reduced (\ref{eq: Bayesian_Spatial_Gaussian_Generic_Marginalized}) to (\ref{eq: Conjugate_Bayesian_LM}) and direct sampling from its posterior is easily achieved as described below (\ref{eq: Conjugate_Bayesian_LM_Posterior}). We will return to the issue of fixing $\{\phi, \delta^2\}$ shortly.

Let us turn to accommodating (\ref{eq: Bayesian_Spatial_Gaussian_Generic}) within (\ref{eq: Conjugate_Bayesian_LM}), which would include directly sampling the spatial random effects $w$ from their marginal posterior $p(w\given y)$. Here, it is instructive to write the joint distribution of $y$ and $w$ in (\ref{eq: Bayesian_Spatial_Gaussian_Generic}) as a linear model,
\begin{equation} \label{eq: Conjugate_Bayesian_LM_Spatial_Effects}
\begin{array}{cccccc}
\underbrace{ \left[ \begin{array}{c} y \\ \mu_\beta \\ 0 \end{array} \right]}
 & = & \underbrace{ \left[ \begin{array}{cc} X & I_n \\ I_p & O \\  O & I_n \end{array} \right] } &
\underbrace{\left[ \begin{array}{c} \beta \\ w \end{array} \right]} & + & \underbrace{ \left[ \begin{array}{c} \eta_1 \\ \eta_2 \\ \eta_3 \end{array} \right]},\\
 y_{*} & = & X_{*} & \gamma & + & \eta
\end{array}\;, %\mbox{ where }\; \eta \sim N(0, \sigma^2V_{y_{*}})\; 
\end{equation} 
where $\eta \sim N(0, \sigma^2V_{y_{*}})$ and $\displaystyle V_{y_{*}} = \begin{bmatrix} \delta^2 I_n & O & O \\ O & V_{\beta} & O\\ O & O & R(\phi)\end{bmatrix}$. If we assume that $\delta^2$ and $\phi$ are fixed at known values, then $V_{y_{*}}$ is fixed. We have a conjugate Bayesian linear regression model $y_{\ast} = X_{\ast}\gamma + \eta$, where $\gamma$ has a flat prior and $\sigma^2\sim IG(a_{\sigma},b_{\sigma})$. Thus,
\begin{align}\label{eq: Conjugate_Bayesian_LM_Spatial_Effects_Posterior}
 p(\gamma, \sigma^2 \given y) &= \underbrace{IG(\sigma^2\given a^*_{\sigma}, b^*_{\sigma})}_{p(\sigma^2\given y)} \times \underbrace{N(\gamma\given M_{*}m_{*}, \sigma^2 M_{*})}_{p(\gamma\given \sigma^2, y)} \; , 
\end{align}
where $a^*_{\sigma} = a_{\sigma} + n/2$, $b^*_{\sigma} = b_{\sigma} + (1/2)\left\{y_{\ast}^{\top}V_{y_{\ast}}^{-1}y_{\ast} - m_{\ast}^{\top}M_{\ast}m_{\ast}\right\}$, $M_{\ast}^{-1} = X_{\ast}^{\top}V_{y_{\ast}}^{-1}X_{\ast}$ and $m_{\ast} = X_{\ast}^{\top}V_{y_{\ast}}^{-1}y_{\ast}$. The posterior mean of $\gamma$ is $\hat{\gamma}= M_{\ast}m_{\ast} = \left(X_{\ast}^{\top}V_{y_{\ast}}^{-1}X_{\ast}\right)^{-1}X_{\ast}^{\top}V_{y_{\ast}}^{-1}y_{\ast}$, which is the generalized least squares estimate obtained from the augmented linear system in (\ref{eq: Conjugate_Bayesian_LM_Spatial_Effects}). Sampling from the posterior proceeds analogous to that described below (\ref{eq: Conjugate_Bayesian_LM_Posterior}). 

From the preceding account we see that fixing the spatial range decay parameter $\phi$ and the noise-to-spatial variance ratio $\delta^2$ casts the Bayesian geostatistical model into a conjugate framework that will allow inference on $\{\beta,w,\sigma^2\}$. Note that multiplying the posterior samples of $\sigma^2$ by the fixed quantity $\delta^2$ fetches us the posterior samples of $\tau^2$. Therefore, we neglect uncertainty in $\phi$ and, partially, for one of the variance components due to fixing their ratio. This, however, provides the computational advantage that inference can be carried out without resorting to expensive iterative algorithms such as MCMC that require several iterations before sampling from the posterior distribution. This computational benefit becomes especially relevant when handling massive spatial data. Furthermore, fixing the values of $\delta^2$ and $\phi$ is not entirely unreasonable given that these parameters are weakly identified by the data \citep[]{zhang2004inconsistent} and difficult to learn from the posterior. Nevertheless, the inference will depend upon these fixed parameters so we discuss a practical approach to fix $\phi$ and $\delta^2$ at reasonable values.

\subsection{Choosing $\phi$ and $\delta^2$}\label{sec: phi_deltasq_fix}
\noindent We can set values for $\phi$ and $\delta^2$ by conducting some simple spatial exploratory data analysis using the ``variogram''. %The variogram for a zero-centered spatial process $w(\ell)$ is defined as
%\begin{equation}\label{eq: variogram}
% \mbox{E}[w(\ell+h) - w(\ell)]^2 = \mbox{var}\left\{w\left(\ell+h\right) - w\left(\ell\right) \right\} = 2\gamma\left(h\right)\;, 
%\end{equation}
%which is meaningful only if the above expression depends solely on $h$ and, whereupon, $\gamma(h)$ is called the ``semivariogram''. If the process $w(\ell)$ is weakly stationary in the sense that the covariance between $w(\ell)$ and $w(\ell')$ is a function only of the separation $h = \ell'-\ell$, then a simple calculation reveals that $\gamma(h) = K_{\theta}(0) - K_{\theta}(h)$, where $K_{\theta}(\ell,\ell') = K_{\theta}(\ell'-\ell) = K(h)$. The variogram is usually computed for the observations $y(\ell)$ or for the residuals from a linear model to ascertain the presence of spatial structure underlying the data after adjusting for explanatory variables. 
Several practical algorithms exist for empirically calculating the variogram (or semivariogram) from observations using finite sample moments. Many of these methods for variograms are now offered in user-friendly \texttt{R} packages hosted by the Comprehensive R Archive Network (CRAN) (\url{https://cran.r-project.org}). As one example, Finley et al. \cite{finley2019efficient} investigate the impact of tree cover and occurrence of forest fires on forest height. They first fit an ordinary linear regression of the form $y_{FH} = \beta_0 + \beta_1 x_{\mbox{tree}} + \beta_2 x_{\mbox{fire}} + \epsilon$ and then compute a variogram for the residuals from the ordinary linear regression. 

\begin{figure}[ht]
\centering
	\includegraphics[scale=0.25]{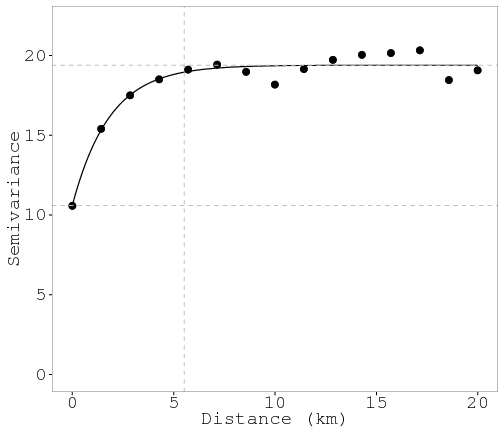}
	\caption{Variogram of the residuals from non-spatial regression indicates strong spatial pattern}\label{fig: variogram_ols_residuals}
\end{figure}   

Figure~\ref{fig: variogram_ols_residuals} depicts the variogram, which informs about the process parameters. The lower horizontal line represents the ``nugget'' or the micro-scale variation captured by the measurement error variance component $\tau^2$. The top horizontal line represents the ``sill'' (or ceiling) which is the total variation captured by $\sigma^2 + \tau^2$. Therefore, the difference between the two horizontal lines is called the ``partial sill'' and is captured by $\sigma^2$. Finally, the vertical line represents the distance beyond which the variogram flattens or the covariance tends to zero. One can provide ``eye-ball'' estimates for these quantities and, in particular, fix the values of $\phi$ and $\delta^2 = \tau^2/\sigma^2$. Fixing these values from the variogram yields the desired highly accessible conjugate framework and the models can be estimated without resorting to Markov chain Monte Carlo (MCMC) as described earlier. Note that instead of $\{\phi,\delta^2\}$, we could also have fixed $\phi$ and any one of the variance components, $\sigma^2$ or $\tau^2$, which would also yield a conjugate model with exact distribution theory. The one slight advantage of fixing $\delta^2$ is that we will get the posterior samples of both $\sigma^2$ and $\tau^2$, the latter obtained simply as $\sigma^2\delta^2$.  

The above crude estimates can be improved using a $K$-fold cross-validation. We split the data randomly into $K$ different folds. Let $S[k]$ be the $k$-th folder of observed points and let $S[-k]$ denote the observed points outside of $S[k]$. For each $k$, we compute the predictive mean $E[y(S[k]) \given y(S[-k])]$. We then compute the ``Root Mean Square Predictive Error'' (RMSPE) \citep{yeniay2002comparison} and choose the value of $\{\phi,\delta^2\}$ corresponding to the smallest RMSPE from a grid of candidate values.  %\textcolor{blue}
The range of the grid is based on interpretation of the hyper-parameters. We suggest a reasonably wide range for $\delta^2$ (e.g., $[0.001, 1000]$), which accommodates one variance component substantially dominating the other in either direction. For the spatial decay $\phi$ we suggest a lower bound of $\frac{3}{\mbox{maximum inter-site distance}}$, which, based upon the exponential covariance function, indicates that the spatial correlation drops below 0.05 at the maximum inter-site distance, and an upper bound that can be initially set as 100 times of the lower bound. Functions like \texttt{variofit} in the R package \texttt{geoR} \citep{geoR} can provide empirical estimates for $\{\phi, \delta^2\}$ from an empirical variogram. After initial fitting, we can shrink the range and refine the grid of the candidate values for more precise estimators.

\subsection{Conjugate Bayesian geostatistical models for massive spatial data}\label{sec: bayes_massive}
\noindent Conjugate models can be estimated by sampling directly from their joint posterior density and, therefore, completely obviate problems associated with MCMC convergence. This is a major computational benefit. However, the challenges in analyzing massive spatial data do not quite end here. When the number of spatial locations providing measurements are in the order of millions as in \cite{finley2019efficient}, then the matrices $K_{\theta}$, $V_{y}$ or $V_{y_{\ast}}$ that we encountered earlier in different model parametrizations will be too massive to be efficiently loaded on to the machine's memory, let alone be computed with. This precludes efficient likelihood computations and has led several researchers to propose models specifically adapted for spatial analysis. We briefly present adaptations of (\ref{eq: Conjugate_Bayesian_LM_Spatial_Effects}) using two different classes of models for massive spatial data: (i) low-rank process models and (ii) NNGP models.   

%In low rank models, the spatial process is approximated as $w(\ell) \approx b_{\theta}^{\top}(\ell)z$, where $b_{\theta}(\ell)$ is an $r\times 1$ vector of $r$ basis functions, each evaluated at $\ell$, and $z$ is an $r\times 1$ vector of coefficients. 
As discussed in Section~\ref{sec: dim_reduction_sparsity}, in low rank models the $n\times 1$ spatial effect $w$ in (\ref{eq: Bayesian_Spatial_Gaussian_Generic}) is replaced by $B_{\theta}z$, where $B_{\theta}$ is the $n\times r$ matrix whose $i$-th row is $b_{\theta}^{\top}(\ell_i)$. Dimension reduction is achieved by fixing $r$ to be much smaller than $n$ so that we only deal with $r$ random effects instead of $n$. The framework in (\ref{eq: Conjugate_Bayesian_LM_Spatial_Effects}) can be easily adapted to this situation as below:
\begin{equation} \label{eq: Conjugate_Bayesian_LM_Spatial_Effects_Low_Rank}
\begin{array}{cccccc}
\underbrace{ \left[ \begin{array}{c} y \\ \mu_\beta \\ 0 \end{array} \right]}
 & = & \underbrace{ \left[ \begin{array}{cc} X & B_{\theta} \\ I_p & O \\  O & I_r \end{array} \right] } &
\underbrace{\left[ \begin{array}{c} \beta \\ z \end{array} \right]} & + & \underbrace{ \left[ \begin{array}{c} \eta_1 \\ \eta_2 \\ \eta_3 \end{array} \right]},\\
 y_{*} & = & X_{*} & \gamma & + & \eta
\end{array}\;, %\mbox{ where }\; \eta \sim N(0, \sigma^2V_{y_{*}})\; 
\end{equation} 
where $\eta \sim N(0, \sigma^2V_{y_{*}})$ and $\displaystyle V_{y_{*}} = \begin{bmatrix} \delta^2 I_n & O & O \\ O & V_{\beta} & O\\ O & O & V_z\end{bmatrix}$ is $(n+p+r)\times (n+p+r)$ and fixed, and $V_z$ is now $r\times r$ instead of the $n\times n$ matrix $R(\phi)$ in (\ref{eq: Conjugate_Bayesian_LM_Spatial_Effects}). Computations for (\ref{eq: Conjugate_Bayesian_LM_Spatial_Effects_Low_Rank}) proceed analogous to those for (\ref{eq: Conjugate_Bayesian_LM_Posterior}), but benefits accrue in terms of storage and the number of floating point operations (flops) when conducting the exact conjugate Bayesian analysis for this model. %The marginal density $p(y_{*}\given \theta, \tau)$ corresponds to the linear model $y_{*} = X_{\ast}\hat{\gamma} + \eta$, where $\hat{\gamma}$ is the generalized least square estimate of $\gamma$ obtained by solving the linear system $X_{\ast}^{\T}V_{y_{*}}^{-1}X_{\ast}\gamma = X_{\ast}^{\T}V_{y_{*}}^{-1}y_{\ast}$. Computational benefits accrue from the block diagonal structure of $V_{y_{\ast}}$.  To be precise, let $V_z^{1/2}$ and $V_{\beta}^{1/2}$ be matrix square roots of $V_z$ and $V_{\beta}$, respectively. For example, $V_{\beta}^{1/2}$ and $V_z^{1/2}$ can be the triangular (upper or lower) Cholesky factor of the $r\times r$ matrices $V_{\beta}$ and $V_{z}$, respectively. Then, the corresponding Cholesky factor of $V_{y_{\ast}}$ is given by by the block diagonal matrix $\displaystyle V_{y_{*}}^{1/2} = \begin{bmatrix} \delta I_n & O & O \\ O & V_{\beta}^{1/2} & O\\ O & O & V_z^{1/2}\end{bmatrix}$. Once we obtain the square root $V_{y_{*}}^{1/2}$, we can make the transformations $\tilde{y}_{\ast} = V_{y_{*}}^{-1/2}y_{\ast}$ and $\tilde{X}_{\ast} = V_{y_{*}}^{-1/2}X_{\ast}$, where $V_{y_{\ast}}^{-1/2}$ is cheaply obtained from $V_{y_{\ast}}^{1/2}$ because it inverts only a triangular matrix. Now the posterior mean $\hat{\gamma}$ can be obtained using ordinary least squares from the model $\tilde{y}_{\ast} = \tilde{X}_{\ast}\hat{\gamma} + e_{\ast}$, where $e_{\ast}\sim N(0, I_{n+p+r})$. %\cite{banerjee2017high} provides a more detailed discussion on hierarchical low-rank models, biases they induce and how bias-adjustments and improvements can be made. 

We now outline the construction of sparse NNGP models. These can be regarded as a special case of Gaussian Markov Random Fields (GMRFs) with a neighborhood structure specified using a directed acyclic graph (DAG). The computational benefits for NNGP models accrue from the ease of inverting sparse matrices. This is immediate from noting that the expense to obtain $V_{y_{\ast}}^{-1}$ in (\ref{eq: Conjugate_Bayesian_LM_Spatial_Effects_Posterior}) is dominated by $R(\phi)^{-1}$. Therefore, if $R(\phi)^{-1}$ is easily available then the inference for $\gamma = \{\beta,w\}$ will be inexpensive. Modeling sparse $R(\phi)^{-1}$ can be easily achieved as follows.
%One approach draws on the concept of sparse precision matrices. There are numerous specifications but the one that is effective, scaleable and easy to compute is based upon modelling the Cholesky decomposition of the precision matrix of $w$ in a sparse manner. To be precise, we assume that $R_{\phi}^{-1} = LDL^{\top}$, where $L$ is unit lower-triangular, i.e., each of its diagonal entries is $1$, with each row containing no more than a few, say $m$, non-zero entries and $D$ is diagonal with positive diagonal entries. The question then is what entries in the lower-triangular part of $L$ should be made zero. A natural thought is to set the $(i,j)$-th entry of $L$ to be zero except when $i>j$ \emph{and} when the spatial location $\ell_i$ is among the $m$ nearest neighbours of $\ell_j$. It remains to discuss choices for the non-zero entries in $L$ and $D$.It turns out that a very effective approximation emerges by letting the non-zero entries in the lower-triangular part of each row of $L$ be obtained by recognising that the lower-triangular elements of $L$ are precisely the coefficients of a linear combination of $w(\ell_j)$'s equating to the conditional expectation $\mbox{E}[w(\ell_i)\given \{w(\ell_j) : j<i\}]$. Consider an $n\times 1$ random vector $w$ distributed as $N(0,K_{\theta})$. 
Writing $N(w\given 0, \sigma^2R_{\phi})$ as $p(w_1)\prod_{i=2}^{n}p(w_i\given w_1, w_2,\ldots, w_{i-1})$ is equivalent to the following set of linear models,
\begin{align*}
 w_1 &= 0 + \eta_1\; \quad \mbox{ and }\; \quad w_i = a_{i1}w_1 + a_{i2}w_2 +  \cdots + a_{i,i-1}w_{i-1} + \eta_i\; \mbox{ for } i=2,\ldots,n\; ,
% \Longrightarrow w &= Aw + \eta;\quad \eta \sim N(0, D)
\end{align*}
or, more compactly, simply $w = Aw + \eta$, where $A$ is $n\times n$ strictly lower-triangular with elements $a_{ij} = 0$ whenever $j \geq i$ and $\eta \sim N(0, D)$ and $D$ is diagonal with diagonal entries $d_{11} = \mbox{var}\{w_1\}$ and $d_{ii} = \mbox{var}\{w_i\given w_j : j < i\}$ for $i=2,\ldots,n$. From the structure of $A$ it is evident that $I-A$ is unit lower-triangular, hence nonsingular, and $R_{\phi} = (I-A)^{-1}D(I-A)^{-\top}$. 

We now introduce sparsity in $R_{\phi}^{-1} = (I-A)^{\top}D(I-A)$ by letting $a_{ij}=0$ whenever $j\geq i$ (since $A$ is strictly lower-triangular) and also whenever $\ell_j$ is not among the $m$ nearest neighbors of $\ell_i$, where $m$ is fixed by the user to be a small number. It turns out that a very effective approximation emerges by recognizing that the lower-triangular elements of $A$ are precisely the coefficients of a linear combination of $w(\ell_j)$'s equating to the conditional expectation $\mbox{E}[w(\ell_i)\given \{w(\ell_j) : j<i\}]$. Thus, the $m\times 1$ vector $\tilde{a}_i$ of non-zero entries in the $i$-th row of $A$ are obtained by solving the $m\times m$ linear system $\tilde{R}_{\phi, N_i, N_i}\tilde{a}_i = R_{\phi, N_i,i}$, where $\tilde{R}_{\phi, N_i, N_i}$ is the $m\times m$ principal submatrix extracted from $R_{\phi}$ corresponding to the $m$ neighbors of $i$ (indexed by elements of a neighbor set $N_i$) and $R_{\phi, N_i,i}$ is the $m\times 1$ vector extracted by choosing the $m$ indices in $N_i$ from the $i$-th column of $R_{\phi}$. Once $\tilde{a}_i$ is obtained, the $i$-th diagonal entry of $D$ is obtained as $d_{ii} = R_{\phi}[i,i] - \tilde{a}_i^{\top}R_{\phi, N_i,i}$. These computations need to be carried out for each $i=2,\ldots,n$ (note that for $i=1$, $d_{11}=\sigma^2$ and $a_{11}=0$), but $m$ can be kept very small (say $5$ or $10$ even if $n~10^7$) so that the expense is $O(nm^3)$ and still feasible. The details can be found in \cite{banerjee2017high}. This notion is familiar in Gaussian Graphical models and have been used in \cite{ve88} and, more recently, in \cite{datta16} and \cite{finley2019efficient} to tackle massive amounts of spatial locations. 

The framework in (\ref{eq: Conjugate_Bayesian_LM_Spatial_Effects}) now assumes the form
\begin{equation} \label{eq: Conjugate_Bayesian_LM_Spatial_Effects_NNGP}
\begin{array}{cccccc}
\underbrace{ \left[ \begin{array}{c} y \\ \mu_\beta \\ 0 \end{array} \right]}
 & = & \underbrace{ \left[ \begin{array}{cc} X & I_n \\ I_p & O \\  O & D^{-1/2}(I-A) \end{array} \right] } &
\underbrace{\left[ \begin{array}{c} \beta \\ w \end{array} \right]} & + & \underbrace{ \left[ \begin{array}{c} \eta_1 \\ \eta_2 \\ \eta_3 \end{array} \right]},\\
 y_{*} & = & X_{*} & \gamma & + & \eta
\end{array}\;, %\mbox{ where }\; \eta \sim N(0, \sigma^2V_{y_{*}})\; 
\end{equation} 
where $\eta \sim N(0, \sigma^2V_{y_{*}})$ and $\displaystyle V_{y_{*}} = \begin{bmatrix} \delta^2 I_n & O & O \\ O & V_{\beta} & O\\ O & O & I_n\end{bmatrix}$ is $(2n+p)\times (2n+p)$ and fixed with much greater sparsity. While this approach can also be subsumed into the framework of (\ref{eq: Conjugate_Bayesian_LM_Spatial_Effects}), its efficient implementation on standard computing architectures needs careful consideration and involves solving a large linear system with $(n+p)\times (n+p)$ coefficient matrix $X_{\ast}^{\top}X_{\ast}$. This matrix is large, but is sparse because of $(I-A)^{\top}D^{-1}(I-A)$. Since $(I-A)$ has at most $m+1$ nonzero entries in each row, an upper bound of nonzero entries in $(I-A)$ is $n(m+1)$ and, therefore, the upper bound in $(I-A)^{\top}D^{-1}(I-A)$ is $n(m+1)^2$. This sparsity can be exploited by sparse linear solvers such as conjugate gradient methods that can be implemented on modest computing environments. 

Sampling from the joint posterior distribution $p(\gamma, \sigma^2\given y_{\ast})$ is achieved in the following manner. First, the least-squares estimate $\hat{\gamma}$ is obtained using a sparse least-square solver using a preconditioned conjugate gradient algorithm. Subsequently, $\sigma^2$ is sampled from its marginal posterior density $IG(a_{\ast}, b_{\ast})$, where $a_{\ast}=a_{\sigma} + n/2$ and $b_{\ast} = b_{\sigma} + (1/2)(y_{\ast} - X_{\ast}\hat{\gamma})^{\top}(y_{\ast} - X_{\ast}\hat{\gamma})$, and we sample one value of $\gamma$ from $N\left(\hat{\gamma}, \sigma^2 \left(X_{\ast}^{\top}V_{y_{\ast}}^{-1}X_{\ast}\right)^{-1}\right)$ using each sampled value of $\sigma^2$. In general, solving $X_{*}^{\top}X_{*} \hat{\gamma} = X_{*}^{\top} y_{*}$ requires $\mathcal{O}(\frac{1}{3}(n+p)^3)$ flops, but when $p \ll n$, the structure of $X_{*}$ and $X_{*}^{\top}X_{*}$ ensures memory requirements in the order of $n(m+1)^2$ and the computational complexity in the order of $nm + n(m+1)^2$ flops.  Details on such implementations on modest computing platforms can be found in \citep{zdb2019}.   

\subsection{Spatial prediction}\label{sec: Spatial_Predictions}
\noindent Let $\tilde{\calL} = \{\tilde{\ell_1}, \tilde{\ell}_2,\ldots,\tilde{\ell}_{\tilde{n}}\}$ be a set of $\tilde{n}$ locations where we wish to predict the outcome $y(\ell)$. Let $\tilde{Y}$ be an $\tilde{n}\times 1$ vector with $i$-th element $\tilde{Y}(\tilde{\ell}_i)$ and let $\tilde{w}$ be the $\tilde{n}\times 1$ vector with elements $w(\tilde{\ell}_i)$. The predictive model augments the joint distribution $p(\theta, w, \beta, \tau, y)$ to
\begin{align}\label{eq: Joint_Model_Prediction}
 p(\theta, \tau, \beta, w, y, \tilde{w}, \tilde{Y}) = p(\theta, \tau, \beta) \times p(w\given \theta) \times p(\tilde{w}\given w, \theta) \times p(y \given \beta, w, \tau)\times  p(\tilde{Y}\given \beta, \tilde{w}, \tau)\;.
\end{align}
The factorization in (\ref{eq: Joint_Model_Prediction}) also implies that $\tilde{Y}$ and $w$ are conditionally independent of each other given $\tilde{w}$ and $\beta$. %The joint posterior distribution of $\{\gamma, \tau^2, \tilde{Y}\}$ is
% \begin{align}\label{eq: Posterior_Joint_Parameter_Predictive}
%  p(\gamma, \tau^2, \tilde{w}, \tilde{Y}\given y) = p(\gamma, \tau^2\given y) \times p(\tilde{w}\given \gamma, \tau^2) \times p(\tilde{Y}\given \tilde{w}, \gamma, \tau^2)\; , 
% \end{align}
%where we have used the conditional independence of $\{\tilde{w}, \tilde{Y}\}$ with $y$ given the other parameters. From (\ref{eq: Posterior_Joint_Parameter_Predictive}), we can see that sampling from the posterior predictive density $p(\tilde{Y}\given y)$ will be achieved by sampling one draw of $\tilde{Y}$ from $p(\tilde{Y}\given \tilde{w}, \gamma, \tau^2)$ for each draw of $\{\gamma, \tau^2\}$ from $p(\gamma, \tau^2\given y)$. 
Predictive inference for spatial data evaluates the posterior predictive distribution $p(\tilde{Y}, \tilde{w}\given y)$. This is the joint posterior distribution for the outcomes and the spatial effects at locations in $\tilde{\calL}$. This distribution is easily derived from (\ref{eq: Joint_Model_Prediction}) as
\begin{align}\label{eq: Joint_Posterior}
 p(\tilde{Y}, \tilde{w}, \beta, w, \theta, \tau \given y) \propto p(\beta, w, \theta, \tau\given y) \times p(\tilde{w}\given w, \theta) \times p(\tilde{Y}\given \beta, \tilde{w}, \tau)\;. 
\end{align}
Sampling from (\ref{eq: Joint_Posterior}) is achieved by first sampling $\{\beta, w, \theta, \tau\}$ from $p(\beta, w, \theta, \tau\given y)$. For each drawn sample, we make one draw of the $\tilde{n}\times 1$ vector $\tilde{w}$ from $p(\tilde{w}\given w, \theta)$ and then, using this sampled $\tilde{w}$, we make one draw of $\tilde{Y}$ from $p(\tilde{Y}\given \beta, \tilde{w}, \tau)$. The resulting samples of $\tilde{w}$ and $\tilde{Y}$ will be draws from the desired posterior predictive distribution $p(\tilde{w}, \tilde{Y}\given y)$. This delivers inference on both the latent spatial random  effect $\tilde{w}$ and the outcome $\tilde{Y}$ at arbitrary locations since ${\cal L}$ can be any finite collection of samples. Summarizing these distributions by computing their sample means, standard errors, and the $2.5$-th and $97.5$-th quantiles (to produce a $95\%$ credible interval) yields point estimates with associated uncertainty quantification. 

It is instructive to see how the entire inference for Gaussian outcomes can be cast into an augmented linear regression model. The predictive model for $\tilde{Y}$ can be written as a spatial regression
\begin{align}\label{eq: Spatial_Regression_Predictive}
 \tilde{Y} = \tilde{X}\beta + \tilde{w} + \tilde{\epsilon}\;;\quad \tilde{w} = Cw + \omega\;,
\end{align}
where $\tilde{X}$ is the $\tilde{n}\times p$ matrix of predictors observed at locations in $\tilde{\calL}$ and $\tilde{\epsilon} \sim N(0, \tilde{D}_{\tau})$, where $\tilde{\epsilon}$ is the $\tilde{n}\times 1$ vector with elements $\epsilon(\tilde{\ell}_i)$. The second equation in (\ref{eq: Spatial_Regression_Predictive}) expresses the relationship between the spatial effects $\tilde{w}$ across the unobserved locations in $\tilde{\calL}$ and the spatial effects across the observed locations in $\calL$. Since there is one underlying random field over the entire domain, the covariance function for the random field specifies the $\tilde{n}\times n$ coefficient matrix $C$. In particular, if $w \sim N(0, K_{\theta})$, then $C = K_{\theta}(\tilde{\calL},\calL)K_{\theta}^{-1}$ and $\omega \sim N(0, F_{\theta})$, where $F_{\theta} = K_{\theta}(\tilde{\calL},\tilde{\calL}) - K_{\theta}(\tilde{\calL},\calL)K_{\theta}^{-1}K_{\theta}(\calL,\tilde{\calL})$.  The model for the data and the predictions is combined into 
\begin{equation} \label{eq: Conjugate_Bayesian_LM_Augmented_Predictive}
\begin{array}{cccccc}
\underbrace{ \left[ \begin{array}{c} y \\ \mu_\beta \\ 0 \\ 0 \\ 0 \end{array} \right]}
 & = & \underbrace{ \left[ \begin{array}{cccc} X & I_n & O & O \\ I_p & O & O & O \\  O & C & - I_{\tilde{n}} & O \\ \tilde{X} & O & I_{\tilde{n}} & -I_{\tilde{n}}\end{array} \right] } &
\underbrace{\left[ \begin{array}{c} \beta \\ w \\ \tilde{w} \\ \tilde{Y} \end{array} \right]} & + & \underbrace{ \left[ \begin{array}{c} \eta_1 \\ \eta_2 \\ \eta_3 \\ \eta_4 \\ \eta_5 \end{array} \right]},\\
 y_{*} & = & X_{*} & \gamma & + & \eta
\end{array}\;, 
\end{equation} 
where $\displaystyle \eta \sim N\left(0, \begin{bmatrix} D_{\tau} & O & O & O & O \\ O & V_{\beta} & O & O & O \\ O & O & K_{\theta} & O & O\\ O & O & O & F_{\theta} & O \\ O & O & O & O & \tilde{D}_{\tau} \end{bmatrix}\right)$. If locations where predictions are sought are fixed by study design, then fitting (\ref{eq: Conjugate_Bayesian_LM_Augmented_Predictive}) using the Bayesian conjugate framework can be beneficial. On the other hand, one can first estimate $\{\beta, w, \sigma^2\}$ and store samples from their posterior distribution. Then, for any arbitrary set of points in $\tilde{\calL}$, for each stored sample of the parameters we draw one sample of $\tilde{w} \sim N(Cw, F_{\theta})$ followed by one draw of $\tilde{Y} \sim N(\tilde{X}\beta +\tilde{w}, \tilde{D}_{\tau})$. The resulting $\{\tilde{w}, \tilde{Y}\}$ will be the desired posterior predictive samples for the latent spatial process and the unobserved outcomes. Again, the advantage of this formulation is that an efficient least squares algorithm to solve (\ref{eq: Conjugate_Bayesian_LM_Augmented_Predictive}) that can exploit the sparsity of the design matrix $X_{\ast}$ will immediately deliver inference on the regression slopes ($\beta$), the spatial process ($w$) at observed points, the interpolated process ($\tildew$) at unobserved points, and the predicted response ($\tilde{Y}$) all at once. 

\section{Illustrative examples}\label{sec: illustrative_examples}
\noindent We present a part of some simulation experiments conducted in \citep{zdb2019}, where we generated data using the spatial regression model in (\ref{eq: BasicModel}) over a set of $n=1200$ spatial locations within a unit square and using an exponential covariance function to specify the spatial process. While $1200$ spatial locations may seem too modest, we use this to draw comparisons with a full GP model that will be too expensive for large datasets. The model included an intercept and a single predictor generated from a standard normal distribution.

We fit a full Gaussian process based model (labeled as full GP in Table~\ref{table:sim}) using the \texttt{spBayes} package in \texttt{R}, a latent NNGP model with $m = 10$ neighbors using the sequential MCMC algorithm described in \cite{datta16} (using the \texttt{spNNGP} package), and the conjugate latent NNGP model described in the preceding section with $m = 10$ neighbors. We will refer to the latent NNGP model fitted using MCMC (with all process parameters unknown) as simply the NNGP or latent NNGP model, while we will explicitly use ``conjugate'' to describe the conjugate latent NNGP model.

These models were trained using $n=1000$ observations, while the remaining $200$ observations were withheld to assess predictive performance. The fixed parameters$\{ \phi, \delta^2 \}$ for the conjugate latent NNGP model were picked through the $K$-fold cross-validation algorithm described in Section~\ref{sec: phi_deltasq_fix}. The intercept and slope parameters in $\beta$ were assigned improper flat priors and an $IG(2,b)$ (mean $b$) prior was used for $\sigma^2$. For the latent NNGP and full GP models, the spatial decay $\phi$ was modeled using a fairly wide uniform prior $U(2.2, 220)$ prior and Inverse-Gamma priors $IG(2,b)$ (mean $b$) were used for the nugget ($\tau^2$) and the partial sill ($\sigma^2$) in order to compare the conjugate Bayesian models with other models. The shape parameter was fixed at $2$ and the scale parameter was set from the empirical estimate provided by the variogram using the \texttt{geoR} package \citep{geoR}. The parameter estimates and performance metrics are provided in Table~\ref{table:sim}. 
\begin{table} 
	\centering
	\def\~{\hphantom{0}}
	\caption{Simulation study summary table: posterior mean (2.5\%, 97.5\%) percentiles} \label{table:sim}
	{\begin{tabular*}{\textwidth}
			{@{}c@{\extracolsep{\fill}}c@{\extracolsep{\fill}} c@{\extracolsep{\fill}}c@{\extracolsep{\fill}}c@{\extracolsep{\fill}}
				c@{\extracolsep{\fill}}c@{\extracolsep{\fill}}@{}}
			\hline
		%	&&\phantom{ab}& \multicolumn{2}{c} {Response} & \phantom{ab}&\multicolumn{2}{c}{Latent} \\
			%[1pt] \cline{4-5} \cline{7-8} \\ [-10pt]
			& True && Full GP  && NNGP & Conj NNGP \\
			\hline
			$\beta_0$ & 1 	&&  1.07(0.72, 1.42) &&  1.10 (0.74, 1.43)  &  1.06 (0.76, 1.46)	  \\
			$\beta_1$ & -5 	&&   -4.97 (-5.02, -4.91) &&  -4.97 (-5.02, -4.91)  &	-4.97 (-5.02, -4.91) \\
			$\sigma^2$ & 2 &&  1.94 (1.63, 2.42) &&  1.95 (1.63, 2.41)  &   1.94 (1.77, 2.12) \\
			$\tau^2$ & 0.2 &&  0.14 (0.07, 0.23)  && 0.15 (0.06, 0.24)   & 0.17 (0.16, 0.19)\\
			$\phi$ & 16 && 19.00 (13.92, 23.66)  &&   18.53 (14.12, 24.17)  &  17.65 \\ 
			\hline
%			KL-D &--&& 1048.1(932.85, 1930.38)   &&  1045.61(634.20, 2321.00) & 795.37(726.18, 876.96) \\ %(mean, 2.5\%, 97.5\%)
			KL-D &--&& 4.45(1.16, 9.95)   &&  5.13(1.66, 11.39) & 3.58(1.27, 8.56) \\ 
			MSE(w) &--&&  297.45(231.62, 444.79 ) && 303.38(228.18, 429.54)  &  313.28 (258.96, 483.75)\\
			RMSPE &--& &  0.94  && 0.94  &  0.94 \\
			time(s) & -- & & 2499 + 23147   &&  109.5   &  12 + 0.6 \\
			\hline
		\end{tabular*}}
		\vspace*{-6pt}
\end{table}
%The summaries for the full Gaussian process based model and the latent NNGP model were based on 1 MCMC chain with $20,000$ iterations. The number of iterations was taken to be large enough to guarantee the convergence of the MCMC chains. We took the first half of the MCMC chains as burn-in. The inference from the conjugate latent NNGP model were based on 300 samples. 300 samples is sufficient for the conjugate latent NNGP model since the conjugate model provides independent samples from the exact posterior distribution. We don't need extra memory for burn-in, and the samples from the conjugate model are more efficient than that from MCMC algorithms.
Table~\ref{table:sim} presents parameter estimates and performance metrics for the candidate models. The inference for $\beta$ is almost indistinguishable across the three models. The full GP and the NNGP fully estimate $\{\sigma^2, \tau^2, \phi \}$ using MCMC and yield very similar results. The conjugate NNGP does not estimate $\phi$ and estimates $\{\sigma^2,\tau^2\}$ subject to the constraint that their ratio $\delta^2$ is fixed. This results, expectedly, in slightly narrower credible intervals for $\sigma^2$ and $\tau^2$. Overall, the parameter estimates are very comparable across the models.

Turning to model comparisons, Zhang et al.\cite{zdb2019} computed the posterior distribution of the Kullback-Leibler divergence (KL-D) by computing it between each candidate model and the full GP for each posterior sample. The KL-D values presented in Table~\ref{table:sim} show no significant differences between the three models in their separation from the true full GP model. The root mean-squared prediction error (RMSPE) values (computed from the hold-out set of 200 locations) across all three models are also similar, further corroborating the comparable predictive performance of the conjugate model with the full Gaussian process. 

In terms of timing (presented in seconds in Table~\ref{table:sim}), the recorded time of the conjugate models includes the time for choosing hyper-parameters through cross-validation and (``+") the time for sampling from the posterior distribution. The recorded time of the full GP model consists of the time for MCMC sampling and (``+") the time for recovering the regression coefficients and predictions. The full latent NNGP model is 200 times faster than the full Gaussian process based model, while the conjugate latent NNGP model uses one tenth of the time required by the latent NNGP model to obtain similar inference on the regression coefficients and latent process. Further simulation experiments conducted by Zhang et al. \citep{zdb2019} also show that interpolation of the latent process is almost indistinguishable between the conjugate and full models. %We defer to details in \citep{zdb2019}.

Next, we present a second simulation example using exactly the same setup as in the preceding example, but with $n=12,000$ spatial locations. Here, we fit a latent NNGP model using the MCMC algorithm in \cite{datta16} and the conjugate latent NNGP model. We used $10,000$ locations for training the models while the remaining $2000$ locations were used for predictive assessment. We summarized the results from the latent NNGP model using a post burn-in posterior sample for $10,000$ iterations. This was deemed adequate based upon the customary convergence diagnostics available in the \texttt{coda} and \texttt{mcse} packages within the \texttt{R} computing environment \citep[][]{coda2006,mcse2011}. The inference from the conjugate latent NNGP model were based on 300 samples. This is sufficient for the conjugate latent NNGP model since the conjugate model provides independent samples from the exact posterior distribution. The full MCMC-based NNGP model took about $1268$ seconds to deliver full Bayesian inference, while the conjugate model took only $99+14=113$ seconds (99 seconds for the cross-validation to fix $\{\phi,\delta^2\}$ and 13 seconds for sampling from the posterior distribution). We found that the RMSPE values for the full latent NNGP and the conjugate model computed using the $2000$ hold-out locations were almost identical (0.67 up to 2 decimal places).       

\begin{figure}
	\centering
	\subfloat[True]{\includegraphics[width = 0.25\linewidth]{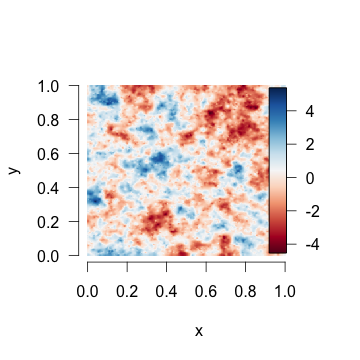}}
	%\subfloat[fullGP]{\includegraphics[width = 0.30\linewidth]{map-w-fullGP.png}}
	\subfloat[NNGP]{\includegraphics[width = 0.25\linewidth]{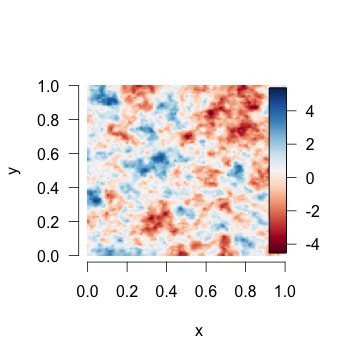}} 
	\subfloat[Conjugate NNGP]{\includegraphics[width = 0.25\linewidth]{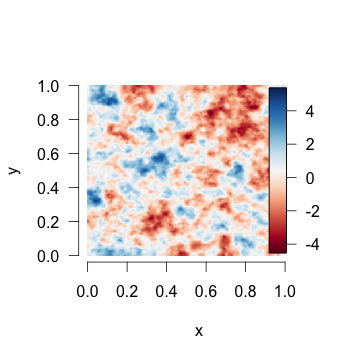}} \\
	\subfloat[CIs of $w$ from NNGP\label{fig: CI_w_LNNGP}]{\includegraphics[width = 0.25\linewidth]{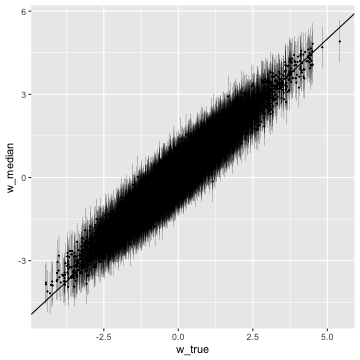}}
	\subfloat[CIs of $w$ from conjugate NNGP \label{fig: CI_w_CLNNGP}]{\includegraphics[width = 0.25\linewidth]{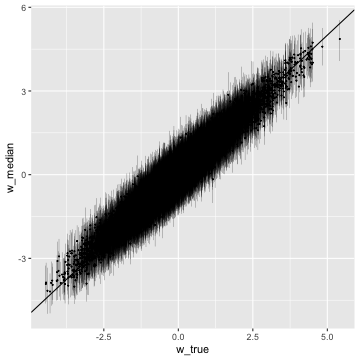}}
	\caption{Interpolated maps of (a) the true generated surface, (b) the posterior means of the spatial latent process $w(s)$ for the NNGP and (c) posterior means of $w(s)$ for the conjugate latent NNGP. The 95\% confidence intervals for the spatial effects $w$ from (d) the NNGP and (e) the conjugate NNGP. The NNGP models were all fit using $m=10$ nearest neighbors. \label{fig:w_com}} 
\end{figure}

The parameter estimates from the full NNGP and conjugate NNGP models in this larger simulation experiment reveal essentially the same story as in Table~\ref{table:sim} so we do not repeat them here. Instead, we focus on the estimation of the latent process and the predictive performance for the two models.  Figure~\ref{fig:w_com} shows interpolated surfaces from the simulation example: \ref{fig:w_com}(a) shows an interpolated map of the ``true'' spatial latent process $w$, while \ref{fig:w_com}(b) and \ref{fig:w_com}(c) present the posterior means of $w(s)$ over the entire domain obtained from the full latent NNGP model and the conjugate latent NNGP model, respectively. 
The recovered spatial residual surfaces are almost indistinguishable, and are comparable to the true interpolated surface of $w(s)$. Figure~\ref{fig:w_com}(d)--(e) present the 95\% credible intervals for the spatial effects $w$ from the latent NNGP model and the conjugate latent NNGP model. These intervals are plotted against the true values of $w$ from the generated model. We found that 9567 out of 10000 credible intervals successfully included the true value for the conjugate model, while the corresponding number was a very comparable 9584 for the full NNGP model.

Turning to a real example, we present a synopsis of the analysis in \cite{zdb2019} of a spatial dataset from NASA comprising sea surface temperature (in degrees Centigrade) observations over 2,827,252 spatial locations of which approximately 90\% (2,544,527) were used for model fitting and the rest were withheld for cross-validatory predictive assessment. Details of the dataset can be found in \url{http://modis-atmos.gsfc.nasa.gov/index.html} and details on the analysis can be found in \cite{zdb2019}. The salient feature of the analysis is that a conjugate Bayesian framework for the NNGP model as in (\ref{eq: Conjugate_Bayesian_LM_Spatial_Effects_NNGP}) was able to deliver full inference including the estimation of the spatial latent effects in about 2387 seconds. Sampling from the posterior distribution was achieved using direct sampling as described below (\ref{eq: Conjugate_Bayesian_LM_Spatial_Effects_NNGP}). Since this algorithm is fast and directly samples from the posterior, hence there is no burn-in period for convergence, it was run over a grid of values of $\{\delta^2,\phi\}$. For each such value, a posterior predictive assessment over the cross-validatory hold-out set was carried out and the value of $\{\delta^2,\phi\}$ producing the least RMSPE  was selected as optimal inputs for which the estimates of $\{\gamma,\sigma^2\}$ were presented.

\begin{figure}[t]
\begin{center}
\subfloat[Posterior mean of sea-surface temperature]{\includegraphics[width=2.75in]{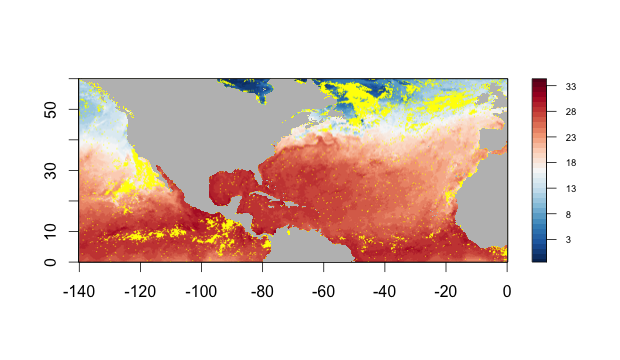}\label{post_mean_SST}}
\subfloat[Posterior predictive mean of latent spatial effects]{\includegraphics[width=2.75in]{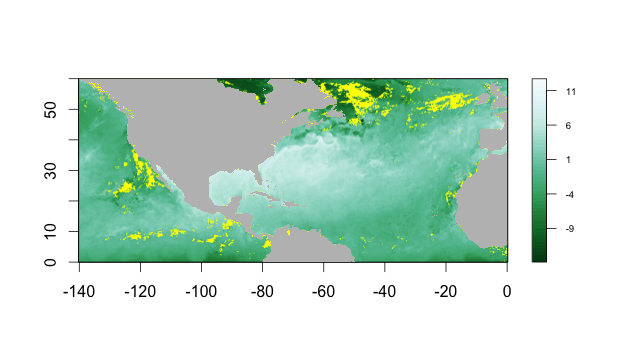}\label{post_mean_w}}
\end{center}
\caption{Posterior predictive maps of sea-surface temperature (in degree centigrade) and latent spatial effects. The land is colored in gray, locations in the ocean without observations are indicated in yellow.} \label{fig: post_mean}
\end{figure}

\section{Spatial Meta-Kriging}\label{sec: meta-kriging}
\noindent %Embedding scalable processes such as the NNGP within a conjugate Bayesian linear regression framework has been shown to enable analysis of spatial data sets in the order of $10^7$ points. While this is impressive, these methods are yet to be tested at ultra-massive scales such as $10^8$ or more points. 
A different approach toward BIG DATA problems relies upon divide and conquer methods. The idea here is divide and conquer (or map and reduce) by pooling posterior inference across a partition of data subsets. Once again consider the Bayesian linear regression model
 \begin{align}\label{Eq: Conjugate_Bayesian_Model}
 p(\beta,\sigma^2\given y) \propto IG(\sigma^2\given a_{\sigma}, b_{\sigma}) \times N(\beta\given \mu_{\beta}, \sigma^2 V_{\beta}) \times N(y\given X\beta, \sigma^2 V_y)\; ,
 \end{align}
where $y$ is $N\times 1$, $X$ is $N\times p$, $\beta$ is $p\times 1$, $V_y$ is a fixed $N\times N$ covariance matrix, $\mu_{\beta}$ is a fixed $p\times 1$ vector and $V_{\beta}$ is a fixed $p\times p$ matrix. The joint posterior density $p(\beta, \sigma^2\given y)$ is available in closed form as
%\begin{align}\label{Eq: Conjugate_Bayesian_Model_Posterior}
 $p(\beta, \sigma^2\given y) = p(\sigma^2\given y) \times p(\beta\given \sigma^2,y)\;,$
%\end{align}
where the marginal posterior density $p(\sigma^2\given y) = IG(\sigma^2\given a^*, b^*)$ and the conditional posterior density $p(\beta\given \sigma^2,y) = N(\beta\given Mm, \sigma^2M)$ with $a^* = a_{\sigma}+N/2$, $b^* = b_{\sigma} + c/2$, $m = V_{\beta}^{-1}\mu_{\beta} + X^{\T}V_y^{-1}y$, $M^{-1} = V_{\beta}^{-1} + X^{\T}V_y^{-1}X$ and $c = \mu_{\beta}^{\T}V_{\beta}^{-1}\mu_{\beta} + y^{\T}V_y^{-1}y - m^{\T}Mm$. Therefore, exact posterior inference can be carried out by first sampling $\sigma^2$ from $IG(a^*, b^*)$ and then sampling $\beta$ from $N(Mm, \sigma^2M)$ for each sampled value of $\sigma^2$. This results in samples from $p(\beta,\sigma^2\given y)$. Besides the fixed hyperparameters in the prior distributions, this exercise requires computing $m$, $M$ and $c$.

Now consider a situation where $N$ is large enough so that memory requirements for computing (\ref{Eq: Conjugate_Bayesian_Model}) is unfeasible. One possible resolution is to replace the likelihood in (\ref{Eq: Conjugate_Bayesian_Model}) with a composite likelihood that assumes independence across blocks formed by partitioning the data into $K$ subsets. We partition the $N\times 1$ vector $y$ into $K$ subvectors with $y_k$ as the $n_k\times 1$ subvector forming the $k$-th subvector, where $\sum_{k=1}^K n_k = N$. The size of the $k$-th subset is $n_k$. These sizes need not be the same across $k$, but will be chosen in a manner so that each of the subsets can be fitted easily with the computational resources available. Also, let $X_k$ be the $n_k\times p$ matrix of predictors corresponding to $y_k$ and let $V_{y_k}$ be the marginal correlation matrix for $y_k$. The conjugate Bayesian model with a block-independent composite likelihood assumes that $y_k = X_k\beta + \epsilon_k$, where $\epsilon_k \stackrel{ind}{\sim} N(0,\sigma^2 V_{y_k})$.
%  \begin{align}\label{Eq: Conjugate_Bayesian_Model_Composite_Likelihood}
%  y_k = X_k\beta + \epsilon_k\;;\;\quad \epsilon_k \stackrel{ind}{\sim} N(0,\sigma^2 D_k)\; .
%  \end{align}
The Bayesian specification is completed by assigning priors to $\sigma^2$ and $\beta$ as in (\ref{Eq: Conjugate_Bayesian_Model}).
%  \begin{align}\label{Eq: Conjugate_Bayesian_Model_Composite_Likelihood}
%  IG(\sigma^2\given a, b) N(\beta\given \mu_{\beta}, \sigma^2 V_{\beta}) \times \prod_{k=1}^K N(y_k\given X_k\beta, \sigma^2 D_k)\; .
%  \end{align}
If we distribute the analysis to $K$ different computing cores, where the $k$-th core fits the above model but only with the likelihood $N(y_k\given X_k\beta, \sigma^2 V_{y_k})$, then the quantities needed for sampling from the full $p(\beta, \sigma^2 \given y)$ can be computed entirely using quantities obtained from the individual subsets of the data. For each $k=1,2,\ldots,K$ we independently compute $m_k = V_{\beta}^{-1}\mu_{\beta} + X_k^{\T}V_{y_k}^{-1}y_k$ and $M_k^{-1} = V_{\beta}^{-1} + X_k^{\T}V_{y_k}^{-1}X_k$ based upon the $k$-th subset of the data. We then combine them to obtain $m = \sum_{k=1}^K (m_k - (1-1/K)V_{\beta}^{-1}\mu_{\beta})$ and $M^{-1} = \sum_{k=1}^K(M_k^{-1} - (1-1/K)V_{\beta}^{-1})$. Subsequently, we compute $c = \mu_{\beta}^{\T}V_{\beta}^{-1}\mu_{\beta} + \sum_{k=1}^Ky_k^{\T}V_{y_k}^{-1}y_k - m^{\T}Mm$. Therefore, sampling from the posterior distribution of $\beta$ and $\sigma^2$ given the entire dataset can be achieved using quantities computed independently from each of the $K$ smaller subsets of the data. There is no need to interact between the subsets and one does not require to store or compute with large objects based upon the entire dataset. This computation can also be done sequentially. We first obtain the posterior distribution $p(\beta,\sigma^2\given y_1)$ based only upon the first data set. This posterior becomes the prior for the next step and we obtain $p(\beta,\sigma^2\given y_1,y_2) \propto p(\beta,\sigma^2\given y_1)\times p (y_2\given \beta,\sigma^2)$ and so on until we arrive at $p(\beta,\sigma^2\given y_1,y_2,\ldots,y_K) \propto p(\beta,\sigma^2\given y_1,y_2,\ldots,y_{K-1})\times p (y_K\given \beta,\sigma^2)$. 

Clearly such exact recovery of the full posterior crucially depends on the conditional independence across the different data blocks (e.g., $p(y_k\given \beta,\sigma^2,y_1,\ldots,y_{k-1}) = p(y_k\given \beta,\sigma^2)$ for each $k=2,\ldots,K$). While this works for uncorrelated outcomes, as in standard linear regression, such recovery is precluded for spatial and spatiotemporal process models and, more generally, for correlated data. Nevertheless, we can develop a general approximation framework for obtaining the full posterior from posterior densities calculated over smaller subsets. One general way to pool information across these individual posteriors is to use the unique \emph{Geometric Median} (GM) of the subset posteriors, as developed by Minsker \citep{Minsker2015}. Assume that the individual posterior densities $p_k \equiv p(\Omega\given y_k)$ reside on a Banach space ${\cal H}$ equipped with norm $\|\cdot\|$. The GM is defined as $\displaystyle \pi^*(\cdot\given y) = \arg\min\limits_{\pi\in\mathcal{H}}\sum_{k=1}^{K}\|p_k-\pi\|_{\rho}$,
% \begin{align}\label{Eq: GM}
% \pi^*(\cdot\given y) = \arg\min\limits_{\pi\in\mathcal{H}}\sum_{k=1}^{K}\|p_k-\pi\|_{\rho}\;  ,
% \end{align}
where $y = (y_1^{\top}, y_2^{\top},\ldots,y_K^{\top})^{\top}$. The norm quantifies the distance between any two posterior densities $\pi_1(\cdot)$ and $\pi_2(\cdot)$ as $\displaystyle \left\|\pi_1 - \pi_2\right\|_{\rho} = \left\|\int \rho(\Omega,\cdot)d(\pi_1-\pi_2)(\Omega)\right\|$, where $\rho(\cdot)$ is a positive-definite kernel function. Assume $\rho(z_1,z_2)=\exp(-\|z_1-z_2\|^2)$. The GM is unique and lies in the convex hull of the individual posteriors, so $\pi^*(\Omega\given y)$ is a legitimate probability density. Specifically, $\pi^*(\Omega\given y)=\sum_{k=1}^{K} \alpha_{\rho,k}(y)p_k$,$\sum_{k=1}^{K}\alpha_{\rho,k}(y)=1$, each $\alpha_{\rho,k}(y)$ being a function of $\rho,y$, so that $\int_{\Omega}\pi^*(\Omega\given y)d\Omega=1$. Computing the GM $\pi^*\equiv \pi^*(\Omega\given y)$ is achieved by an iterative algorithm that estimates $\alpha_{\rho,k}(y)$ from the subset posteriors $p_k$ for each $k=1,2,\ldots,K$. To further elucidate, we use a well known result that the GM $\pi^*$ satisfies $\pi^*=\frac{\sum_{k=1}^{K}\|p_k-\pi^*\|_{\rho}^{-1}p_k}{\sum_{k=1}^{K}\|p_k-\pi^*\|_{\rho}^{-1}}$, so that $\alpha_{\rho,k}(y)=\frac{\|p_k-\pi^*\|_{\rho}^{-1}}{\sum_{j=1}^{K}\|p_k-\pi^*\|_{\rho}^{-1}}$.
There is no apparent closed-form solution for $\alpha_{\rho,k}(y)$ satisfying this equation, so Weiszfeld's algorithm \citep{Minsker2015} is used to estimate these functions.

This approach has been extended to spatial process settings by Guhaniyogi and Banerjee \citep{GB2018, GB2019}. The advantage here is that one can use existing Bayesian geostatistical software to sample from the posterior distributions of the different subsets. This can be performed either in parallel over multiple cores or across different machines altogether. One then needs to save only the post burn-in samples and execute Weiszfeld's algorithm to these samples. Weiszfeld's algorithm is extremely fast and easy to program. 

\section{Discussion}\label{sec: discussion}
\noindent This article has attempted to provide a brief overview of how some Bayesian geostatistical models designed for large spatial and/or spatiotemporal datasets can be further scaled up to analyze massive datasets with observed locations in the order of $10^6$ or more by exploiting the familiar theory of conjugate Bayesian linear regression models and adapting them to incorporate latent spatial processes. The resulting distribution theory is available in closed form, thereby circumventing the need for iterative algorithms such as MCMC or INLA. We have also provided a brief overview of a distributed approach (spatial meta-kriging) that relies upon analyzing exclusive subsets of the data and combining them to approximate the full posterior in the spirit of a spatial meta-analysis.  

Of course, this requires some compromise in terms of full Bayesian inference. Some parameters need to be provided as fixed inputs for the distribution theory to be available in closed form. Learning about these input parameters will be done using exploratory data analysis and cross-validation methods. A practical approach that seems to be quite effective for analyzing massive datasets in modest computing environments is to choose the optimal value of the process parameters based upon the minimum RMSPE over hold-out locations. While such approaches may produce slightly shrunk credible and prediction intervals due to the effect of fixing a parameter, the effect is seen to be moderate in practical spatial analysis and the approach could form a useful tool for quick spatial analysis within the Bayesian paradigm for massive spatial datasets. However, the method of learning about these parameters is still ad-hoc and can possibly be improved with more sophisticated optimization methods. Nevertheless, the approach outlined here can be a useful tool in the spatial analyst's toolbox for exploring Bayesian spatial regression at massive scales. We also point out that the conjugate Bayesian linear regression framework can accommodate almost all of the model-based GP approximations for dimension reduction or sparsity induction. Any spatial covariance structure that leads to efficient computations can, in principle, be used. 

While the article has focused on the NNGP as a choice for introducing sparsity in the model, more general GMRF specifications are also admissible here. In fact, there has been much recent activity within the framework of Vecchia approximations \citep[see, e.g.,][]{katzfuss2017general,katzfuss2018predictions}, where the models are being derived using DAGs over the expanded set of observations and process realizations. While certainly promising, their benefits and improvements over GMRFs are yet to be demonstrated in large scale case studies. For Vecchia type of likelihoods, there is also interest in choosing the number of neighbors. First, it should be intuitively clear that DAGs constructed using shrunk neighbor sets will yield probability models farther away from the full model as the neighbor sets get smaller. To see this, consider a random vector $w = (w_{A}^{\top}, w_{B}^{\top})^{\top}$, where $A$ and $B$ are mutually exclusive sets containing indices for the elements of $w$, and let $p(w) = p(w_{A})p(w_{B}\given w_{A})$ denote the joint probability density for $w$. Consider two submodels $p_1(w) = p(w_{A})\times p(w_{B}\given w_{N_{1B}})$ and $p_2(w) = p(w_{A})\times p(w_{B}\given w_{N_{2B}})$, where $N_{2B}\subset N_{1B} \subset A$. The model $p_2$ will be farther than $p_1$ from $p$ in the terms of the Kullback-Leibler divergence:
{\small
\begin{equation}\label{eq: KL_divergence_proof}
 \begin{split}
  KL(p\|\|p_2) - KL(p\|\|p_1) &= \int \left\{\log\left(\frac{p(w)}{p_2(w)}\right) - \log\left(\frac{p(w)}{p_1(w)}\right)\right\}p(w)dw \\
  &= \int \log\left(\frac{p_1(w)}{p_2(w)}\right)p(w)dw = \int \log\left(\frac{p(w_{B}\given w_{N_{1B}})}{p(w_{B}\given w_{N_{2B}})}\right)p(w)dw \\
%  &= \int \log\left(\frac{p(w_{B}\given w_{N_{1B}})}{p(w_{B}\given w_{N_{2B}})}\right)p(w_B,w_{N_1B},w_{A\setminus N_{1B}})dw \\
  &= \int \log\left(\frac{p(w_{B}\given w_{N_{1B}})}{p(w_{B}\given w_{N_{2B}})}\right)p(w_B\given w_{N_{1B}})p(w_{N_{1B}})dw_{B}dw_{N_{1B}} \\
  &= \int \left\{\int\log\left(\frac{p(w_{B}\given w_{N_{1B}})}{p(w_{B}\given w_{N_{2B}})}\right)p(w_B\given w_{N_{1B}})dw_{B} \right\}p(w_{N_{1B}})dw_{N_{1B}} \geq 0\;,
 \end{split}
\end{equation}
}
where we have used the fact that $A\setminus N_{1B}$ is mutually exclusive of $N_{1B}$ and, crucially, also of $N_{2B}$ (since $N_{2B}\subset N_{1B}$) to legitimately integrate out $w_{A\setminus N_{1B}}$. The final conclusion follows from a customary application of Jensen's inequality to show that the inner integral in the last equation is non-negative. Equation~(\ref{eq: KL_divergence_proof}) provides an alternate distribution-free proof of a result for Gaussian likelihoods by Guinness (Theorem~1 in \cite{guinness18}). These results also indicate that the ordering of the variables to construct the approximation can affect model performance and certain designs to determine the ordering can produce improved results (as demonstrated in \cite{guinness18}). Datta et al. \cite{datta16b} argued against fixing the neighborhoods in spatiotemporal contexts (since neighbors in space and neighbors in time may not align) and demonstrate a computationally efficient method to learn about neighbors in spatiotemporal domains. 

Finally, we point toward a few future directions of research in this domain. Much of the spatial literature on modeling massive spatial data have focused upon scalability of models and algorithms. There is still work to be done on evaluating the inferential performance of these models at such massive scales. How important is uncertainty quantification at such scales? How do GP based approaches compare with deep learning with neural networks in spatial analysis? Another area where the cross-validatory learning approaches for process hyperparameters will struggle is in multivariate contexts, where the number of hyperparameters is higher than here. These are some areas of research where we believe the statistical community still has much to offer.

\bibliographystyle{ba}
\bibliography{Banerjee}

\begin{thebibliography}{55}
\newcommand{\enquote}[1]{``#1''}
\expandafter\ifx\csname natexlab\endcsname\relax\def\natexlab#1{#1}\fi
\expandafter\ifx\csname url\endcsname\relax
  \def\url#1{{\tt #1}}\fi
\expandafter\ifx\csname urlprefix\endcsname\relax\def\urlprefix{URL }\fi
\ifx\endbibitem\undefined \let\endbibitem\relax\fi

\bibitem[{Abdulah et~al.(2018)Abdulah, Ltaief, Sun, Genton, and
  Keyes}]{abdulah2018}
Abdulah, S., Ltaief, H., Sun, Y., Genton, M., and Keyes, D. (2018).
\newblock \enquote{ExaGeoStat: A high performance unified software for
  geostatistics on manycore systems.}
\newblock {\em IEEE Transactions on Parallel and Distributed Systems\/}, 29:
  2771--2784.
\endbibitem

\bibitem[{Banerjee(2017)}]{banerjee2017high}
Banerjee, S. (2017).
\newblock \enquote{High-Dimensional Bayesian Geostatistics.}
\newblock {\em Bayesian Analysis\/}, 12: 583--614.
\endbibitem

\bibitem[{Banerjee et~al.(2014)Banerjee, Carlin, and
  Gelfand}]{banerjee2014hierarchical}
Banerjee, S., Carlin, B.~P., and Gelfand, A.~E. (2014).
\newblock {\em Hierarchical modeling and analysis for spatial data\/}.
\newblock CRC Press, Boca Raton, FL.
\endbibitem

\bibitem[{Banerjee et~al.(2010)Banerjee, Finley, Waldmann, and
  Ericcson}]{ban10}
Banerjee, S., Finley, A.~O., Waldmann, P., and Ericcson, T. (2010).
\newblock \enquote{Hierarchical Spatial Process Models for Multiple Traits in
  Large Genetic Trials.}
\newblock {\em Journal of the American Statistical Association\/}, 105:
  506--521.
\endbibitem

\bibitem[{Banerjee et~al.(2008)Banerjee, Gelfand, Finley, and Sang}]{ban08}
Banerjee, S., Gelfand, A.~E., Finley, A.~O., and Sang, H. (2008).
\newblock \enquote{Gaussian Predictive Process Models for Large Spatial
  Datasets.}
\newblock {\em Journal of the Royal Statistical Society, Series B\/}, 70:
  825--848.
\endbibitem

\bibitem[{Cressie(1993)}]{cres93}
Cressie, N. (1993).
\newblock {\em Statistics for Spatial Data\/}.
\newblock Wiley-Interscience, revised edition.
\endbibitem

\bibitem[{Cressie and Johannesson(2008)}]{cres08}
Cressie, N. and Johannesson, G. (2008).
\newblock \enquote{Fixed Rank Kriging for Very Large Data Sets.}
\newblock {\em Journal of the Royal Statistical society, Series B\/}, 70:
  209--226.
\endbibitem

\bibitem[{Cressie et~al.(2010)Cressie, Shi, and Kang}]{cres10}
Cressie, N., Shi, T., and Kang, E.~L. (2010).
\newblock \enquote{Fixed Rank Filtering for Spatio-temporal Data.}
\newblock {\em Journal of Computational and Graphical Statistics\/}, 19:
  724--745.
\endbibitem

\bibitem[{Cressie and Wikle(2011)}]{creswikle11}
Cressie, N. A.~C. and Wikle, C.~K. (2011).
\newblock {\em Statistics for Spatio-temporal Data\/}.
\newblock Wiley series in probability and statistics. Hoboken, N.J. Wiley.
\newline\urlprefix\url{http://opac.inria.fr/record=b1133266}
\endbibitem

\bibitem[{Datta et~al.(2016a)Datta, Banerjee, Finley, and Gelfand}]{datta16}
Datta, A., Banerjee, S., Finley, A.~O., and Gelfand, A.~E. (2016a).
\newblock \enquote{Hierarchical Nearest-Neighbor Gaussian Process Models for
  Large Geostatistical Datasets.}
\newblock {\em Journal of the American Statistical Association\/}, 111:
  800--812.
\newline\urlprefix\url{http://dx.doi.org/10.1080/01621459.2015.1044091}
\endbibitem

\bibitem[{Datta et~al.(2016b)Datta, Banerjee, Finley, Hamm, and
  Schaap}]{datta16b}
Datta, A., Banerjee, S., Finley, A.~O., Hamm, N. A.~S., and Schaap, M. (2016b).
\newblock \enquote{Non-separable Dynamic Nearest-Neighbor Gaussian Process
  Models for Large spatio-temporal Data With an Application to Particulate
  Matter Analysis.}
\newblock {\em Annals of Applied Statistics\/}, 10: 1286--1316.
\newline\urlprefix\url{http://dx.doi.org/10.1214/16-AOAS931}
\endbibitem

\bibitem[{Du et~al.(2009)Du, Zhang, and Mandrekar}]{du09}
Du, J., Zhang, H., and Mandrekar, V.~S. (2009).
\newblock \enquote{Fixed-domain Asymptotic Properties of Tapered Maximum
  Likelihood Estimators.}
\newblock {\em Annals of Statistics\/}, 37: 3330--3361.
\endbibitem

\bibitem[{Finley et~al.(2015)Finley, Banerjee, and Gelfand}]{finbangel15}
Finley, A.~O., Banerjee, S., and Gelfand, A.~E. (2015).
\newblock \enquote{{spBayes} for Large Univariate and Multivariate
  Point-Referenced Spatio-Temporal Data Models.}
\newblock {\em Journal of Statistical Software\/}, 63(13): 1--28.
\newline\urlprefix\url{http://www.jstatsoft.org/v63/i13/}
\endbibitem

\bibitem[{Finley et~al.(2019)Finley, Datta, Cook, Morton, Andersen, and
  Banerjee}]{finley2019efficient}
Finley, A.~O., Datta, A., Cook, B.~C., Morton, D.~C., Andersen, H.~E., and
  Banerjee, S. (2019).
\newblock \enquote{Efficient algorithms for Bayesian Nearest Neighbor Gaussian
  Processes.}
\newblock {\em Journal of Computational and Graphical Statistics\/}, 28(2):
  401--414.
\endbibitem

\bibitem[{Flegal and Jones(2011)}]{mcse2011}
Flegal, J. and Jones, G. (2011).
\newblock \enquote{Implementing Markov chain Monte Carlo: Estimating with
  confidence.}
\newblock In Brooks, S., Gelman, A., Jones, G., and Meng, X. (eds.), {\em
  Handbook of Markov Chain Monte Carlo\/}, 175--197. Chapman and Hall/CRC
  Press, Boca Raton, FL.
\endbibitem

\bibitem[{Furrer et~al.(2006)Furrer, Genton, and Nychka}]{fur06}
Furrer, R., Genton, M.~G., and Nychka, D. (2006).
\newblock \enquote{Covariance Tapering for Interpolation of Large Spatial
  Datasets.}
\newblock {\em Journal of Computational and Graphical Statistics\/}, 15:
  503--523.
\endbibitem

\bibitem[{Gelfand et~al.(2010)Gelfand, Diggle, Fuentes, and
  Guttorp}]{geldigfuegut}
Gelfand, A., Diggle, P., Fuentes, M., and Guttorp, P. (2010).
\newblock {\em Handbook of Spatial Statistics\/}.
\newblock Boca Raton, FL: CRC Press.
\endbibitem

\bibitem[{Gelman et~al.(2013)Gelman, Carlin, Stern, Dunson, Vehtari, and
  Rubin}]{gelman2013}
Gelman, A., Carlin, J.~B., Stern, H.~S., Dunson, D.~B., Vehtari, A., and Rubin,
  D.~B. (2013).
\newblock {\em Bayesian Data Analysis, 3rd Edition\/}.
\newblock Chapman \& Hall/CRC Texts in Statistical Science. Chapman \&
  Hall/CRC.
\endbibitem

\bibitem[{Gneiting and Guttorp(2010)}]{gnei10}
Gneiting, T. and Guttorp, P. (2010).
\newblock \enquote{Continuous-parameter Spatio-temporal Processes.}
\newblock In Gelfand, A., Diggle, P., Fuentes, M., and Guttorp, P. (eds.), {\em
  Handbook of Spatial Statistics\/}, 427--436. CRC Press, Boca Raton, FL.
\endbibitem

\bibitem[{Guhaniyogi and Banerjee(2018)}]{GB2018}
Guhaniyogi, R. and Banerjee, S. (2018).
\newblock \enquote{Meta-Kriging: Scalable Bayesian Modeling and Inference for
  Massive Spatial Datasets.}
\newblock {\em Technometrics\/}, 60(4): 430--444.
\newline\urlprefix\url{https://doi.org/10.1080/00401706.2018.1437474}
\endbibitem

\bibitem[{Guhaniyogi and Banerjee(2019)}]{GB2019}
--- (2019).
\newblock \enquote{Multivariate spatial meta-kriging.}
\newblock {\em Statistics and Probability Letters\/}, 144: 3--8.
\newline\urlprefix\url{https://doi.org/10.1080/00401706.2018.1437474}
\endbibitem

\bibitem[{Guinness(2018)}]{guinness18}
Guinness, J. (2018).
\newblock \enquote{Permutation and Grouping Methods for Sharpening Gaussian
  Process Approximations.}
\newblock {\em Technometrics\/}, 60(4): 415--429.
\newline\urlprefix\url{https://doi.org/10.1080/00401706.2018.1437476}
\endbibitem

\bibitem[{Heaton et~al.(2019)Heaton, Datta, Finley, Furrer, Guinness,
  Guhaniyogi, Gerber, Gramacy, Hammerling, Katzfuss, Lindgren, Nychka, Sun, and
  Zammit-Mangion}]{heatoncontest2019}
Heaton, M., Datta, A., Finley, A., Furrer, R., Guinness, J., Guhaniyogi, R.,
  Gerber, F., Gramacy, R., Hammerling, D., Katzfuss, M., Lindgren, F., Nychka,
  D., Sun, F., and Zammit-Mangion, A. (2019).
\newblock \enquote{Methods for Analyzing Large Spatial Data: A Review and
  Comparison.}
\newblock {\em Journal of Agricultural, Biological and Environmental
  Statistics\/}, 24(3): 398--425.
\newline\urlprefix\url{https://doi.org/10.1007/s13253-018-00348-w}
\endbibitem

\bibitem[{Huang and Sun(2018)}]{huangsun2018}
Huang, H. and Sun, Y. (2018).
\newblock \enquote{Hierarchical low-rank approximation of likelihoods for large
  spatial datasets.}
\newblock {\em Journal of Computational and Graphical Statistics\/}, 27:
  110--118.
\endbibitem

\bibitem[{Katzfuss(2013)}]{katzfuss2013}
Katzfuss, M. (2013).
\newblock \enquote{Bayesian nonstationary modeling for very large spatial
  datasets.}
\newblock {\em Environmetrics\/}, 24: 189--200.
\endbibitem

\bibitem[{Katzfuss(2017)}]{katzfussmultires}
--- (2017).
\newblock \enquote{A multi-resolution approximation for massive spatial
  datasets.}
\newblock {\em Journal of the American Statistical Association\/}, 112:
  201--214.
\newline\urlprefix\url{http://dx.doi.org/10.1080/01621459.2015.1123632}
\endbibitem

\bibitem[{Katzfuss and Cressie(2012)}]{katz12}
Katzfuss, M. and Cressie, N. (2012).
\newblock \enquote{Bayesian hierarchical spatio-temporal smoothing for very
  large datasets.}
\newblock {\em Environmetrics\/}, 23: 94--107.
\endbibitem

\bibitem[{Katzfuss and Guinness(2017)}]{katzfuss2017general}
Katzfuss, M. and Guinness, J. (2017).
\newblock \enquote{A General Framework for Vecchia Approximations of Gaussian
  Processes.}
\newblock {\em arXiv preprint arXiv:1708.06302\/}.
\endbibitem

\bibitem[{Katzfuss et~al.(2018)Katzfuss, Guinness, Gong, and
  Zilber}]{katzfuss2018predictions}
Katzfuss, M., Guinness, J., Gong, W., and Zilber, D. (2018).
\newblock \enquote{Vecchia approximations of Gaussian-process predictions.}
\newblock {\em arXiv preprint arXiv:1805.03309\/}.
\endbibitem

\bibitem[{Kaufman et~al.(2008)Kaufman, Scheverish, and Nychka}]{kauf08}
Kaufman, C.~G., Scheverish, M.~J., and Nychka, D.~W. (2008).
\newblock \enquote{Covariance Tapering for Likelihood-Based Estimation in Large
  Spatial Data Sets.}
\newblock {\em Journal of the American Statistical Association\/}, 103:
  1545--1555.
\endbibitem

\bibitem[{Lindgren et~al.(2011)Lindgren, Rue, and
  Lindstrom}]{lindgrenruelindstrom2011}
Lindgren, F., Rue, H., and Lindstrom, J. (2011).
\newblock \enquote{An explicit link between Gaussian fields and Gaussian Markov
  random fields: the stochastic partial differential equation approach.}
\newblock {\em Journal of the Royal Statistical Society: Series B (Statistical
  Methodology)\/}, 73(4): 423--498.
\newline\urlprefix\url{http://dx.doi.org/10.1111/j.1467-9868.2011.00777.x}
\endbibitem

\bibitem[{Ma and Kang(2017)}]{Ma2017FusedGP}
Ma, P. and Kang, E.~L. (2017).
\newblock \enquote{Fused Gaussian Process for Very Large Spatial Data.}
\newblock {\em arXiv:1702.08797v3\/}.
\endbibitem

\bibitem[{Minsker(2015)}]{Minsker2015}
Minsker, S. (2015).
\newblock \enquote{Geometric median and robust estimation in banach spaces.}
\newblock {\em Bernoulli\/}, 21: 2308--2335.
\endbibitem

\bibitem[{Moller and Waagepetersen(2003)}]{moll03}
Moller, J. and Waagepetersen, R.~P. (2003).
\newblock {\em Statistical Inference and Simulation for Spatial Point
  Processes\/}.
\newblock Chapman and Hall, first edition.
\endbibitem

\bibitem[{Nychka et~al.(2015)Nychka, Bandyopadhyay, Hammerling, Lindgren, and
  Sain}]{nychka2015}
Nychka, D., Bandyopadhyay, S., Hammerling, D., Lindgren, F., and Sain, S.
  (2015).
\newblock \enquote{A Multiresolution Gaussian Process Model for the Analysis of
  Large Spatial Datasets.}
\newblock {\em Journal of Computational and Graphical Statistics\/}, 24(2):
  579--599.
\newline\urlprefix\url{http://dx.doi.org/10.1080/10618600.2014.914946}
\endbibitem

\bibitem[{Nychka et~al.(2002)Nychka, Wikle, and
  Royle}]{Nychka_Wikle_Royle_2002}
Nychka, D., Wikle, C., and Royle, J.~A. (2002).
\newblock \enquote{{Multiresolution models for nonstationary spatial covariance
  functions}.}
\newblock {\em Statistical Modelling\/}, 2(4): 315--331.
\endbibitem

\bibitem[{Plummer et~al.(2006)Plummer, Best, Cowles, and Vines}]{coda2006}
Plummer, M., Best, N., Cowles, K., and Vines, K. (2006).
\newblock \enquote{CODA: Convergence Diagnosis and Output Analysis for MCMC.}
\newblock {\em R News\/}, 6(1): 7--11.
\newline\urlprefix\url{https://journal.r-project.org/archive/}
\endbibitem

\bibitem[{Rasmussen and Williams(2005)}]{rasm08}
Rasmussen, C.~E. and Williams, C. K.~I. (2005).
\newblock {\em Gaussian Processes for Machine Learning\/}.
\newblock Cambridge, MA: The MIT Press, first edition.
\endbibitem

\bibitem[{{Ribeiro Jr} and Diggle(2012)}]{geoR}
{Ribeiro Jr}, P.~J. and Diggle, P.~J. (2012).
\newblock {\em geo{R}: a package for geostatistical analysis\/}.
\newblock R package version 1.7-4.
\newline\urlprefix\url{https://cran.r-project.org/web/packages/geoR}
\endbibitem

\bibitem[{Rua and Held(2005)}]{rueheld04}
Rua, H. and Held, L. (2005).
\newblock {\em Gaussian Markov Random Fields : Theory and Applications\/}.
\newblock Monographs on statistics and applied probability. Chapman and
  Hall/CRC Press, Boca Raton, FL.
\newline\urlprefix\url{http://opac.inria.fr/record=b1119989}
\endbibitem

\bibitem[{Rue et~al.(2009)Rue, Martino, and Chopin}]{ruemartinochopin2009}
Rue, H., Martino, S., and Chopin, N. (2009).
\newblock \enquote{Approximate Bayesian inference for latent Gaussian models by
  using integrated nested Laplace approximations.}
\newblock {\em Journal of the Royal Statistical Society: Series B (Statistical
  Methodology)\/}, 71(2): 319--392.
\newline\urlprefix\url{http://dx.doi.org/10.1111/j.1467-9868.2008.00700.x}
\endbibitem

\bibitem[{Sang and Huang(2012)}]{sang12}
Sang, H. and Huang, J.~Z. (2012).
\newblock \enquote{A Full Scale Approximation of Covariance Functions for Large
  Spatial Data Sets.}
\newblock {\em Journal of the Royal Statistical society, Series B\/}, 74:
  111--132.
\endbibitem

\bibitem[{Sang et~al.(2011)Sang, Jun, and Huang}]{sang11}
Sang, H., Jun, M., and Huang, J. (2011).
\newblock \enquote{Covariance approximation for large multivariate spatial
  datasets with an application to multiple climate model errors.}
\newblock {\em Annals of Applied Statistics\/}, 4: 2519--2548.
\endbibitem

\bibitem[{Schabenberger and Gotway(2004)}]{scha04}
Schabenberger, O. and Gotway, C.~A. (2004).
\newblock {\em Statistical Methods for Spatial Data Analysis\/}.
\newblock Chapman and Hall/CRC Press, Boca Raton, FL, first edition.
\endbibitem

\bibitem[{Shi et~al.(2017)Shi, Kang, Konomi, Vemaganti, and
  Madireddy}]{Shi2017}
Shi, H., Kang, E.~L., Konomi, B.~A., Vemaganti, K., and Madireddy, S. (2017).
\newblock \enquote{Uncertainty Quantification Using the Nearest Neighbor
  Gaussian Process.}
\newblock In Chen, D.-G., Jin, Z., Li, G., Li, Y., Liu, A., and Zhao, Y.
  (eds.), {\em New Advances in Statistics and Data Science\/}, 89--107. Cham,
  Switzerland: Springer International Publishing.
\newline\urlprefix\url{https://doi.org/10.1007/978-3-319-69416-0_6}
\endbibitem

\bibitem[{Shi and Cressie(2007)}]{shicres07}
Shi, T. and Cressie, N. (2007).
\newblock \enquote{Global Statistical Analysis of MISR Aerosol Data: A Massive
  Data Product From NASA's Terra Satellite.}
\newblock {\em Environmetrics\/}, 18: 665--680.
\endbibitem

\bibitem[{Stein(1999)}]{stein99}
Stein, M.~L. (1999).
\newblock {\em Interpolation of Spatial Data: Some Theory for Kriging\/}.
\newblock Springer, first edition.
\endbibitem

\bibitem[{Stein et~al.(2004)Stein, Chi, and Welty}]{stein04}
Stein, M.~L., Chi, Z., and Welty, L.~J. (2004).
\newblock \enquote{Approximating Likelihoods for Large Spatial Data Sets.}
\newblock {\em Journal of the Royal Statistical society, Series B\/}, 66:
  275--296.
\endbibitem

\bibitem[{Stroud et~al.(2017)Stroud, Stein, and Lysen}]{stroud17}
Stroud, J., Stein, M.~L., and Lysen, S. (2017).
\newblock \enquote{Bayesian and Maximum Likelihood Estimation for Gaussian
  Processes on an Incomplete Lattice.}
\newblock {\em Journal of Computational and Graphical Statistics\/}, 26:
  108--120.
\newline\urlprefix\url{http://dx.doi.org/10.1080/10618600.2016.1152970}
\endbibitem

\bibitem[{Vecchia(1988)}]{ve88}
Vecchia, A.~V. (1988).
\newblock \enquote{Estimation and Model Identification for Continuous Spatial
  Processes.}
\newblock {\em Journal of the Royal Statistical society, Series B\/}, 50:
  297--312.
\endbibitem

\bibitem[{Wikle and Cressie(1999)}]{wikle99}
Wikle, C. and Cressie, N. (1999).
\newblock \enquote{A dimension reduced approach to space-time Kalman
  filtering.}
\newblock {\em Biometrika\/}, 86: 815--829.
\endbibitem

\bibitem[{Wikle(2010)}]{wikle_2011}
Wikle, C.~K. (2010).
\newblock \enquote{Low-Rank Representations for Spatial Processes.}
\newblock {\em Handbook of Spatial Statistics\/}, 107--118.
\newblock Gelfand, A. E., Diggle, P., Fuentes, M. and Guttorp, P., editors,
  Chapman and Hall/CRC, pp. 107-118.
\endbibitem

\bibitem[{Yeniay and Goktas(2002)}]{yeniay2002comparison}
Yeniay, O. and Goktas, A. (2002).
\newblock \enquote{A comparison of partial least squares regression with other
  prediction methods.}
\newblock {\em Hacettepe Journal of Mathematics and Statistics\/}, 31(99):
  99--101.
\endbibitem

\bibitem[{Zhang(2004)}]{zhang2004inconsistent}
Zhang, H. (2004).
\newblock \enquote{Inconsistent estimation and asymptotically equal
  interpolations in model-based geostatistics.}
\newblock {\em Journal of the American Statistical Association\/}, 99(465):
  250--261.
\endbibitem

\bibitem[{Zhang et~al.(2019)Zhang, Datta, and Banerjee}]{zdb2019}
Zhang, L., Datta, A., and Banerjee, S. (2019).
\newblock \enquote{Practical Bayesian Modeling and Inference for Massive
  Spatial Datasets On Modest Computing Environments.}
\newblock {\em Statistical Analysis and Data Mining: The ASA Data Science
  Journal\/}, 12(3): 197--209.
\newline\urlprefix\url{https://doi.org/10.1002/sam.11413}
\endbibitem

\end{thebibliography}

\end{document}